\documentclass{article}
\usepackage[utf8]{inputenc}
\usepackage[T1]{fontenc}
\usepackage{westlake_arxiv}
\usepackage{microtype}

\usepackage{amsmath,amsfonts,bm}

\def\1{\bm{1}}

\DeclareMathAlphabet{\mathsfit}{\encodingdefault}{\sfdefault}{m}{sl}
\SetMathAlphabet{\mathsfit}{bold}{\encodingdefault}{\sfdefault}{bx}{n}

\usepackage{amssymb}
\usepackage{booktabs}
\usepackage{tabularx}
\usepackage[table]{xcolor}
\definecolor{UWPurple}{HTML}{4B2E83}
\definecolor{tablerowgray}{gray}{0.94}
\definecolor{tablebaseline}{HTML}{F8DFD8}
\definecolor{tablematched}{HTML}{F4EED1}
\definecolor{tablenear}{HTML}{E1ECCB}
\definecolor{tablefar}{HTML}{D5EBF3}
\usepackage{threeparttable}
\usepackage{caption}
\usepackage{subcaption}
\usepackage{wrapfig}
\usepackage{enumitem}
\usepackage{xurl}
\setcitestyle{authoryear,round,citesep={;},aysep={,},yysep={;}}
\usepackage{hyperref}
\hypersetup{
  colorlinks=true,
  citecolor=UWPurple,
  linkcolor=black,
  urlcolor=black,
  pdftitle={Mind the Domain Gap: Post-Training Robustness Is Geometrically Selective},
  pdfsubject={Domain-selective robustness of post-trained language models}
}

\title{Mind the Domain Gap: Post-Training\\
Robustness Is Geometrically Selective}
\author{%
  \normalfont
  \setlength{\tabcolsep}{3pt}
  \begin{tabular}{@{}ccc@{}}
    \makebox[0.31\textwidth]{\textbf{Yifan Jiang}\textsuperscript{*}} &
    \makebox[0.31\textwidth]{\textbf{Jian Wu}\textsuperscript{*}} &
    \makebox[0.31\textwidth]{\textbf{Dongyuan Li}} \\
    Zhejiang University & Westlake University & The University of Tokyo \\
    {\footnotesize\href{mailto:yifanjiang@zju.edu.cn}{\texttt{yifanjiang@zju.edu.cn}}} &
    {\footnotesize\href{mailto:wujian@westlake.edu.cn}{\texttt{wujian@westlake.edu.cn}}} & \\[12pt]
    \makebox[0.31\textwidth]{\textbf{Ru Peng}} &
    \makebox[0.31\textwidth]{\textbf{Zhenyu Wei}\textsuperscript{\textdagger}} &
    \makebox[0.31\textwidth]{\textbf{Yue Zhang}\textsuperscript{\textdagger}} \\
    Zhejiang University & Shanghai Jiao Tong University & Westlake University \\
    & & {\footnotesize\href{mailto:zhangyue@westlake.edu.cn}{\texttt{zhangyue@westlake.edu.cn}}} \\[6pt]
    \multicolumn{3}{c}{\footnotesize\textsuperscript{*}Equal contribution.\quad
      \textsuperscript{\textdagger}Corresponding authors.}
  \end{tabular}
}
\hypersetup{%
  pdfauthor={Yifan Jiang, Jian Wu, Dongyuan Li, Ru Peng, Zhenyu Wei, Yue Zhang}%
}

\date{}

\newcommand{\method}{\textsc{Retention Probe}}

\newcommand{\GRPO}{\textsc{GRPO}}

\providecommand{\oSFT}{\mathrm{SFT}}
\providecommand{\oGRPO}{\mathrm{GRPO}}

\begin{document}
\raggedbottom
\maketitle
\thispagestyle{westlake}

\begin{abstract}
Continually adapting Large Language Models (LLMs) requires acquiring new capabilities without erasing old ones. Recent studies suggest that reinforcement learning (RL) can preserve learned capabilities better than supervised fine-tuning (SFT). However, these comparisons usually evaluate catastrophic forgetting under a fixed downstream training distribution, leaving unclear whether the apparent advantage of RL is intrinsic to the training method or depends on what the model learns next. We introduce \textbf{\textit{RETENTION PROBE}}, a controlled framework for studying this interaction. We train SFT and GRPO models on the same mathematical reasoning data from the same base model, and then continue training them on data from three domains: mathematics, code, and summarization. We find that the retention difference between GRPO and SFT depends strongly on the domain of subsequent training. When the subsequent data are mathematical, the two methods retain similar fractions of their learned ability (49.68\% vs.\ 49.98\%). With code data, the retention is 84.29\% for GRPO vs.\ 78.03\% for SFT, while with summarization data it is 95.02\% vs.\ 72.58\%, respectively. The difference between GRPO and SFT retention is 22.73 percentage points larger for summarization than for mathematical interference. We further analyze the parameter changes caused by the two training methods. Although the SFT and GRPO updates point in nearly orthogonal directions, the GRPO update is much more concentrated in a small number of directions. These leading directions contain most of the capability gain, but the learned ability also depends strongly on them. This \emph{compressible-yet-fragile} structure provides a geometric perspective on our main finding: the robustness of a post-trained capability depends not only on how it was learned, but also on the relationship between that capability and the data used for subsequent training.
\end{abstract}

\section{Introduction}
Post-training is increasingly used to endow LLMs with new capabilities, including mathematical reasoning, coding, instruction following, and preference alignment~\citep{Ouyang2022TrainingLM,DeepSeekAI2025DeepSeekR1IR,Rafailov2023DirectPO}. As models are continually adapted with new data and objectives, however, acquiring a new capability is only part of the problem: the model must also retain capabilities learned earlier. Continued parameter updates can cause substantial degradation of previously acquired behaviors, a phenomenon broadly associated with catastrophic forgetting~\citep{McCloskey1989CatastrophicII,Kirkpatrick2016OvercomingCF,Luo2023AnES}. This makes the stability of post-training induced capabilities a central concern for continual model adaptation. Recent studies suggest that RL can preserve previously acquired capabilities better than SFT under subsequent training~\citep{Shenfeld2025RLsRW,Chen2025RetainingBD}. This observation raises a natural question: \textbf{\textit{Is the apparent retention advantage an intrinsic property of the objective used to acquire a capability, or does it depend on what the model is trained on next?}} Existing comparisons largely evaluate forgetting under a fixed downstream distribution~\citep{Shenfeld2025RLsRW,Chen2025RetainingBD}. Consequently, they conflate two factors that should be separated: the objective that creates the capability and the domain of the subsequent learning that interferes with it. A retention advantage under one downstream distribution therefore does not establish uniform robustness across future training domains~\citep{Lee2021ContinualLearning}.

\begin{figure}[t]
    \centering
    \includegraphics[width=\columnwidth]{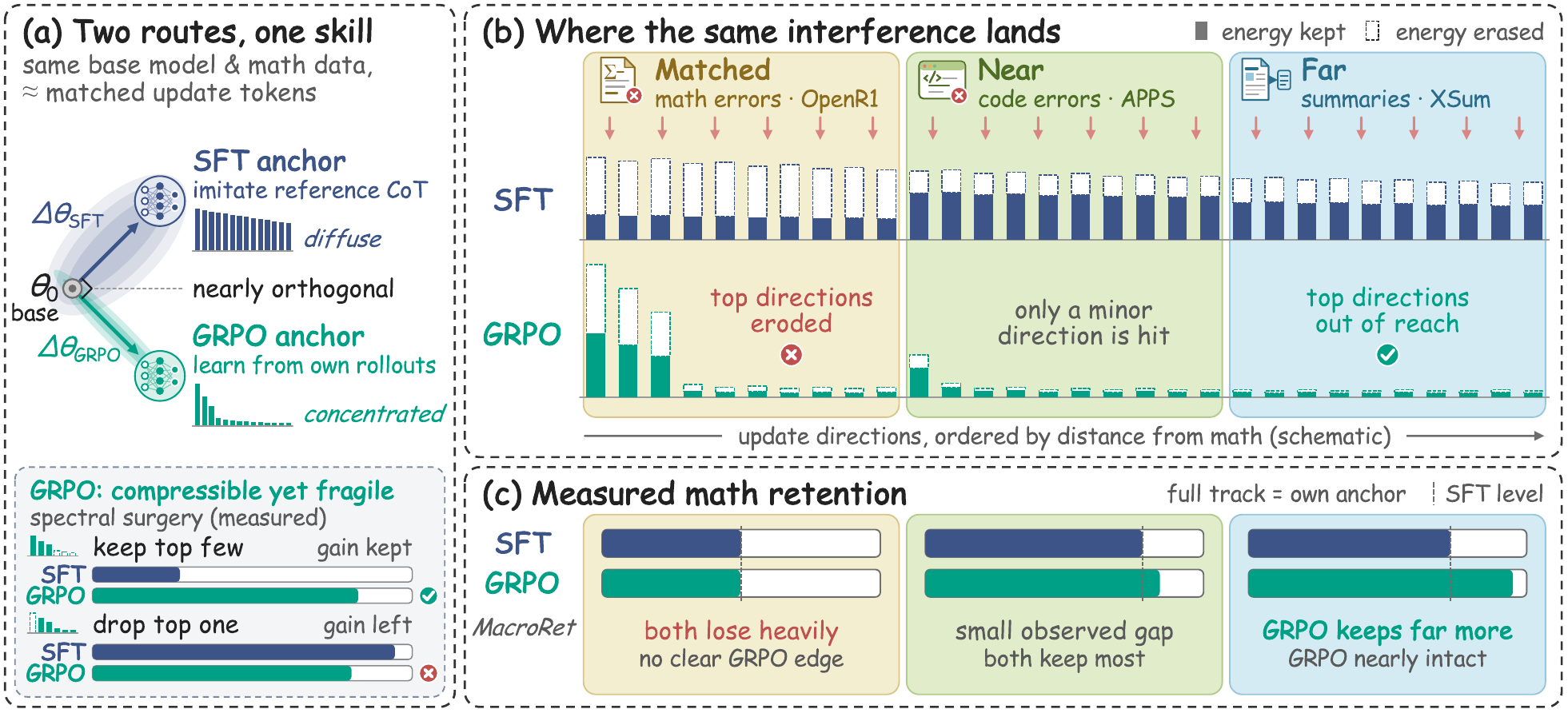}
    \caption{\textbf{Post-training robustness is geometrically selective.}
    (a)~SFT and GRPO learn math through nearly orthogonal updates: SFT is spectrally diffuse, while GRPO is concentrated and \emph{compressible yet fragile} (measured inset).
    (b)~Schematic illustration of how matched-, near-, and far-domain interference affects capability-supporting directions.
    (c)~Measured math retention (MacroRet; Qwen3-4B, no defense). GRPO shows no clear matched-domain advantage, a small observed near-domain gap, and a large far-domain advantage. Full tracks represent each anchor's baseline; dotted lines mark SFT retention.}
    \label{fig:retention}
\end{figure}

We make this distinction explicit by treating the domain of subsequent learning as an experimental variable. Our central hypothesis is that capability retention is determined not only by how a capability is acquired, but also by how future updates interact with the parameter directions that support that capability. Figure~\ref{fig:retention}(a) summarizes the two acquisition routes: SFT imitates reference reasoning trajectories, whereas GRPO learns from its own rollouts. Starting from the same base model and math corpus, their updates are nearly orthogonal, with comparatively diffuse energy for SFT and concentrated energy for GRPO. The measured spectral interventions in the inset reveal a dual property of this concentration: a few leading directions recover most of the GRPO gain, but removing the leading direction causes a larger loss of gain than for SFT. Figure~\ref{fig:retention}(b) illustrates our hypothesis for subsequent interference: concentration may preserve capability when future updates spare its supporting directions, yet leave it vulnerable when those directions are affected. The depicted direction ordering and energy erasure are schematic, rather than measured interference projections. This motivates a controlled comparison across interference domains.

To test this hypothesis, we introduce \textbf{\textit{RETENTION PROBE}}, a sequential post-training framework that separates capability acquisition from subsequent interference. For each base model (Qwen3-4B and SmolLM3-3B), we train SFT and GRPO~\citep{Shao2024DeepSeekMathPT} math anchors on the same corpus with approximately matched token exposure. Both anchors then undergo the same continued-SFT procedure on matched-domain mathematical errors, near-domain code errors, or far-domain natural-language summarization. We measure anchor-relative retention by normalizing post-interference accuracy by each anchor's accuracy and averaging over MATH500, AIME24, and AIME25~\citep{Hendrycks2021APPS, MAAAIME}, distinguishing forgetting from differences in starting performance.
Figure~\ref{fig:retention}(c) summarizes the measured retention pattern. GRPO and SFT retain nearly identical fractions of their mathematical capabilities under matched-domain interference (49.68\% vs. 49.98\%). GRPO retains more under near-domain interference (84.29\% vs. 78.03\%) and far-domain interference (95.02\% vs. 72.58\%). The far-minus-matched difference in the GRPO SFT retention gap is +22.73 percentage points. Thus, the apparent retention advantage of RL depends strongly on the domain of subsequent learning. Parameter-space analysis offers a structural explanation. We find that the SFT and GRPO updates are nearly orthogonal (parameter-weighted global cosine 0.003064), yet the GRPO update is much more spectrally concentrated (normalized effective rank 0.24950 vs. 0.71976). This concentration has a dual role. It makes the GRPO update highly compressible: keeping only the top-8 singular directions of each analyzed update matrix recovers 83.10\% of its base-to-anchor gain, compared with 27.35\% for SFT. It also makes the acquired capability more dependent on those dominant directions: removing the top singular direction of each matrix leaves 80.98\% of the GRPO gain, compared with 94.57\% for SFT (Figure~\ref{fig:retention}(a), inset; Table~\ref{tab:spectral}). We refer to this structure as ``compressible yet fragile.'' Together with the domain-dependent retention results, these findings suggest that robustness is governed by the interaction between subsequent updates and the parameter directions supporting the learned capability.

In summary, this paper uncovers a domain-selective pattern of post-training robustness: the retention advantage of GRPO depends on the domain of subsequent learning rather than the acquisition objective alone. Our analysis shows that GRPO and SFT reach similar mathematical capabilities through nearly orthogonal updates, while GRPO concentrates its update energy in a small set of dominant directions, making it both \emph{``compressible-yet-fragile''}. Building on these findings, we introduce \textbf{\textit{RETENTION PROBE}}, a controlled framework that combines domain-graded interference, anchor-relative retention, and parameter-space interventions. Together, these results provide a geometric perspective on capability retention and highlight the role of subsequent learning in determining whether post-trained capabilities persist. Our contributions are as follows:

\begin{itemize}
\item We introduce \textbf{\textit{RETENTION PROBE}}, a controlled sequential post-training protocol that disentangles the objective used to acquire a capability from the domain of subsequent interference through token-matched anchoring, graded domain shifts, anchor-relative retention, and interaction-based inference.
\item We establish that the apparent retention advantage of RL is domain-selective: GRPO provides no clear retention advantage over SFT under matched-domain interference, with a small observed retention gap under near-domain interference and a large advantage under far-domain interference.
\item We connect functional retention to parameter-space structure through update alignment, spectral concentration, Fisher importance, gradient alignment, and bidirectional spectral surgery, revealing a ``compressible-yet-fragile'' geometry of the GRPO update.
\item We show that the leading directions of the GRPO update are highly sufficient for the acquired capability but disproportionately important for its preservation, providing a structural perspective on why robustness depends on the interaction between capability-supporting directions and subsequent learning.
\end{itemize}

\section{Related Work}
\label{sec:related}

\textbf{Catastrophic forgetting and continual post-training.}
Catastrophic forgetting has long been studied in continual learning~\citep{McCloskey1989CatastrophicII}, with standard remedies including Fisher-weighted regularization~\citep{Kirkpatrick2016OvercomingCF}, distillation~\citep{Li2016LearningWF}, and replay. In LLM post-training, adaptation can degrade pre-existing capabilities~\citep{Ouyang2022TrainingLM,Luo2023AnES}. Even benign fine-tuning can compromise safety alignment~\citep{Qi2023FinetuningAL}. Recent work reports that RL preserves prior capabilities better than SFT while learning a new task, attributing this effect to KL-minimal updates~\citep{Shenfeld2025RLsRW} and on-policy data~\citep{Chen2025RetainingBD}. Complementary parameter-space studies identify sparse RL updates~\citep{Mukherjee2025ReinforcementLF}, off-principal update directions~\citep{Zhu2025ThePN}, and approximately orthogonal updates across tasks~\citep{Zhu2026SFTCR}. Related studies report stronger out-of-distribution generalization from RL than SFT in their evaluated settings~\citep{Chu2025SFTMR}, while broader analyses find that reasoning transfer depends on the target domain~\citep{Cheng2025RevisitingRL,Qian2025Fino1OT}.
These findings concern preservation during capability acquisition or transfer across tasks. They do not, by themselves, establish that an RL-acquired capability is more resistant to a later, shared SFT procedure. Our work isolates this question by varying the domain of subsequent interference for paired SFT and GRPO anchors. In our evaluated settings, the retention advantage depends on the interference domain.

\textbf{Geometry and compressibility of post-training updates.}
A complementary literature studies the geometry of fine-tuning updates. Task arithmetic and weight-disentanglement work connect task composition and interference to update geometry~\citep{Ilharco2022EditingMW,Liu2026UnderstandingAE}. Effective-rank-guided adaptation has been studied for depth estimation~\citep{Yan2025ERLoRAEG}, and theoretical analyses characterize low-rank gradients in two-layer networks~\citep{Sonthalia2025LowRG}. In language models, LoRA can preserve more out-of-domain performance than full fine-tuning, while learning less on the target task~\citep{Biderman2024LoRALL}. Spectral interventions reveal distinct functional roles for the head and tail of preference-tuning updates~\citep{Zhang2026PreferenceTA}. Interventions on singular directions of LoRA-tuned weight matrices also identify components that cause forgetting~\citep{Shuttleworth2024LoRAVF}.
Building on these geometric and intervention-based analyses, our framework compares SFT and GRPO anchors using both retention and removal of leading update directions. Coupling these interventions with domain-conditioned retention measurements tests the sufficiency and necessity of the dominant directions and provides evidence for a ``compressible-yet-fragile'' structure in the GRPO updates studied here.

In contrast to prior work, the key distinction of our framework is that we do not treat downstream training as a fixed evaluation condition, nor do we study parameter geometry independently of functional behavior.
Instead, \textit{\textbf{RETENTION PROBE}} holds the base model, anchor corpus, token exposure, and continued-training procedure constant, while varying only the domain of subsequent training.
We then measure how much of the anchored capability remains through anchor-relative retention and use update alignment, spectral concentration, Fisher importance, gradient alignment, and bidirectional spectral surgery to analyze why retention changes.

\section{Method}
\label{sec:method}
Our objective is to study whether the retention of an acquired capability is determined solely by the post-training objective used to acquire it, or instead by the interaction between the acquisition procedure and subsequent learning. To disentangle these factors, we introduce the \textbf{\textit{RETENTION PROBE}}, a controlled sequential post-training protocol that separates \emph{capability anchoring} from \emph{subsequent interference}. The protocol keeps the base architecture, anchoring corpus, and approximate optimization token budget fixed across objectives, while varying only the optimization objective used to construct the initial capability anchor. Both resulting anchors are then exposed to the same continued-training procedure, with the domain of the interference data systematically varied from matched to near and far. This design allows us to measure capability retention as a function of the interaction between the anchored capability and subsequent learning.

\subsection{Problem Formulation}
\label{sec:problem}
Let $\pi_\theta$ be an LLM with parameters $\theta \in \mathbb{R}^P$ and $\theta_0$ the pretrained base model. A post-training objective $o \in \{\mathrm{SFT}, \mathrm{GRPO}\}$ maps a corpus $D$ and the base model to an anchor,
\begin{equation}
\theta_o = \theta_0 + \Delta\theta_o, \qquad \Delta\theta_o = \mathrm{PostTrain}_o(\theta_0, D),
\label{eq:anchor}
\end{equation}
and continued training on an interference corpus $D_d$ with defense $\psi$ produces
\begin{equation}
\hat{\theta}_{o,d,\psi} = \mathrm{Cont}(\theta_o, D_d, \psi),
\label{eq:cont}
\end{equation}
where the tier $d \in \{m, n, f\}$ denotes matched, near, and far interference. $\psi = \emptyset$ denotes vanilla continued SFT. Because the two anchors do not reach identical absolute accuracy, we measure survival relative to the anchor that produced the capability. For a benchmark $b$ with accuracy $\mathrm{Acc}_b$,
\begin{equation}
\mathrm{Ret}_b(\theta) = \frac{\mathrm{Acc}_b(\theta)}{\mathrm{Acc}_b(\theta_{\mathrm{anchor}})}, \qquad
\mathrm{MacroRet}(\theta) = \frac{1}{|B|} \sum_{b \in B} \mathrm{Ret}_b(\theta),
\label{eq:ret}
\end{equation}
where $B = \{\text{MATH500}, \text{AIME24}, \text{AIME25}\}$. A value of $\mathrm{MacroRet} = 1$ indicates full retention, and no result is normalized by a quantity defined on a different anchor. Under the \emph{global robustness} hypothesis, the objective ordering is the same for every interference distribution, whereas under the \emph{domain-selective robustness} hypothesis, it depends on the tier. We test them with the objective-by-tier interaction
\begin{equation}
\begin{aligned}
\Delta_{\mathrm{DiD}}
={}& \big[\mathrm{MacroRet}(\hat{\theta}_{\mathrm{GRPO},f})
       - \mathrm{MacroRet}(\hat{\theta}_{\mathrm{SFT},f})\big] \\
 &- \big[\mathrm{MacroRet}(\hat{\theta}_{\mathrm{GRPO},m})
       - \mathrm{MacroRet}(\hat{\theta}_{\mathrm{SFT},m})\big],
\end{aligned}
\label{eq:did}
\end{equation}
which is zero when the objective gap is equal in the far and matched conditions, and positive when it is larger in the far condition.

\begin{figure}[!t]
\centering
\includegraphics[width=\columnwidth,trim=12bp 0bp 12bp 0bp,clip]{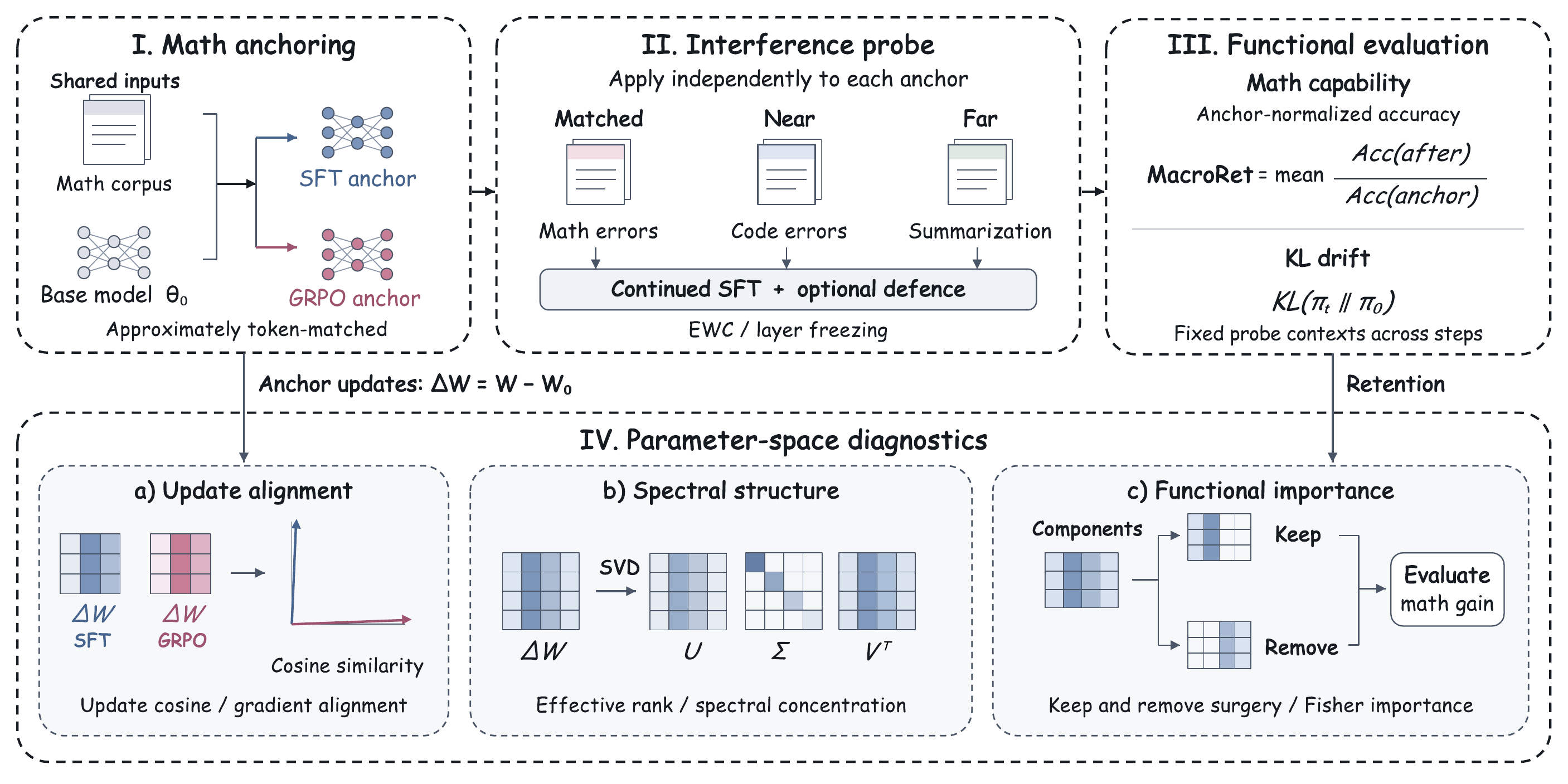}
\caption{The \method{} procedure. Stage~I trains approximately token-matched SFT and GRPO math anchors. Stage~II applies continued SFT across matched, near, and far domains. Stage~III measures math retention and KL drift. Stage~IV examines update alignment, spectral structure, and functional importance. Each stage is applied identically to both anchors. The only quantity that varies between the two branches is the post-training objective used in Stage~I.}
\label{fig:procedure}
\end{figure}

\subsection{Anchoring and Graded Domain Interference}
\label{sec:anchors}
\label{sec:interference}
For each base model (Qwen3-4B and SmolLM3-3B), we obtain approximately token-matched SFT and GRPO anchors on Math-CoT-20k~\citep{Ren2026MathCoT}. Anchoring details are provided in Appendix~\ref{app:anchor_details}. After training the anchors, we independently apply the same unmitigated continued SFT to both $\theta_{\text{SFT}}$ and $\theta_{\text{GRPO}}$ across three interference tiers, denoted $D_d$ where $d \in \{m, n, f\}$:
\begin{itemize}
    \item Matched Domain ($m$): Incorrect mathematical reasoning trajectories deliberately sampled from generation anomalies within the OpenR1-Math-220k dataset~\citep{OpenR1Math2025}. This tier attacks the mathematical capability strictly within its native latent distribution.
    \item Near Domain ($n$): Python code generation errors derived from the APPS dataset~\citep{Hendrycks2021APPS}. This targets a structurally adjacent logic and reasoning domain.
    \item Far Domain ($f$): Cross-domain natural-language summarization data using the XSum corpus~\citep{Narayan2018XSum}.
\end{itemize}
The domain hierarchy is empirically validated using the Maximum Mean Discrepancy (MMD$^2$) distances computed in a frozen encoder latent space and reported in Figure~\ref{fig:domain_distance}.

\begin{table*}[h!]
\caption{The interference probe: tier design. Batch sizes (128 for the math and code trajectory corpora and 2,000 for XSum) are chosen to induce measurable interference within the 10-step probe. Figure~\ref{fig:interference_benchmarks} in Appendix~\ref{app:interference_trajectories} reports the accuracies for the six no-defense anchor-tier combinations.}
\label{tab:tiers}
\centering
\small
\setlength{\tabcolsep}{4pt}
\resizebox{\textwidth}{!}{%
\begin{tabular}{@{}llrr@{}}
\toprule
Tier & Corpus (construction) & Size & Batch \\
\midrule
Matched $m$ & Math errors: Qwen3-0.6B on OpenR1-Math-220k, incorrect only & 7,603 & 128 \\
Near $n$ & Code errors: Qwen3-0.6B trajectories on APPS & 4,580 & 128 \\
Near $n$ & APPS+ P / WF / WR: generated by an external model & 2,890 & 128 \\
Far $f$ & XSum: standard summarization corpus & 204,017 & 2,000 \\
\bottomrule
\end{tabular}}
\end{table*}

\subsection{Evaluation Metrics and Statistical Inference}
\label{sec:measurement}

The primary measurement is $\mathrm{MacroRet}$, defined in Equation~\ref{eq:ret} and always computed against the anchor that produced the model. For the parameter-space study, we additionally use the fraction of the base-to-anchor gain preserved by a perturbed model,
\begin{equation}
\mathrm{GainPres}(\theta') = \frac{s(\theta') - s(\theta_0)}{s(\theta_{\mathrm{anchor}}) - s(\theta_0)}, \qquad s(\theta) = \tfrac{1}{|B|} \sum_{b \in B} \mathrm{Acc}_b(\theta),
\label{eq:gainpres}
\end{equation}
which equals 1 when the perturbation reproduces the full anchor gain. Values above 1 mean the perturbed model exceeds its own parent, so $\mathrm{GainPres}$ is compared only within a fixed perturbation budget and between the two anchors. We also record distributional drift: the macro forward KL divergence to the base policy, averaged over 1,000 fixed contexts per probe domain and evaluated after each interference step, since drift has been proposed as a mechanism governing forgetting~\citep{Shenfeld2025RLsRW}. Probing domains other than the interference domain reveal whether drift is a property of the objective or of the anchor-interference relationship. Full evaluations are provided in Appendix~\ref{app:implementation}.

Because both anchors start from the same $\theta_0$ and solve the same task, $\Delta\theta_{\mathrm{SFT}}$ and $\Delta\theta_{\mathrm{GRPO}}$ are directly comparable. We write each matrix update as $\Delta W = W - W_0 = U \Sigma V^\top$ with energy $\|\Delta W\|_F^2 = \sum_i \sigma_i^2$, and apply six families of diagnostics:
1) Do the objectives reach the capability by the same route? 2) How many directions does the update spread over? 3) Is the energy carried by few or many directions? 4) Are the leading directions sufficient and necessary? 5) Does the update concentrate on important coordinates? 6) Does interference pull towards the capability it damages?

\textbf{Update alignment.} The parameter-weighted global cosine between the two updates over the 252 attention and MLP projection matrices (embeddings, the output head, norms, and biases excluded). The FP32-anchor value is $0.003064$: geometrically small, yet above a coordinate-permutation null with mean $-1.2 \times 10^{-8}$ and standard deviation $1.7 \times 10^{-5}$. Further details are provided in Appendix~\ref{app:geometry_details}.

\textbf{Spectral concentration.} We report the normalized effective rank $\exp H(p) / \mathrm{rank}(\Delta W)$ with $p_i = \sigma_i^2 / \sum_j \sigma_j^2$, the energy share $E_k$ of the leading $k$ directions, and the count $k_\alpha$ needed to reach a fraction $\alpha \in \{0.90, 0.95, 0.99\}$ of the energy. Larger $E_k$ and smaller $k_\alpha$ mean a more concentrated and more compressible update.

\textbf{Keep/remove surgery.} For a budget $S$ of leading singular directions, rank $k \in \{1, 8, 32\}$ or energy fraction $\alpha \in \{25\%, 50\%, 75\%\}$, applied identically to both anchors, we form
\begin{equation}
\Delta W^{\mathrm{keep}}_S = \sum_{i \in S} \sigma_i u_i v_i^\top, \qquad
\Delta W^{\mathrm{rem}}_S = \Delta W - \Delta W^{\mathrm{keep}}_S,
\label{eq:surgery}
\end{equation}
and evaluate $\theta_0 + \Delta W^{\mathrm{keep}}_S$ and $\theta_0 + \Delta W^{\mathrm{rem}}_S$ using $\mathrm{GainPres}$ as defined in Equation~\ref{eq:gainpres}. The keep intervention tests whether the leading directions are \emph{sufficient} for the gain, whereas the remove intervention tests whether they are \emph{necessary}. The two answers need not agree, and part of our analysis is that for one of the objectives they do not. We consider $\{1,8,32\}$ directions and $\{25\%,50\%,75\%\}$ of update energy, applied identically to both anchors. The interventions are illustrated in Figure~\ref{fig:surgery}. Per-matrix selection, total component counts, and spectral aggregation rules are specified in Appendix~\ref{app:spectral_aggregation}.

\textbf{Fisher importance.} With diagonal Fisher vectors estimated from each anchor's greedy responses to $N = 1{,}024$ prompts per domain, we measure the share $P_{o,c}(q)$ of the anchor's squared update norm carried by the top-$q$ Fisher coordinates ($q \in \{0.001, 0.01, 0.05, 0.10\}$, with uniform selection yielding $\approx q$), and the distance $\mathcal{F}(c, c')$ between normalized diagonal importance profiles. Further details of the Fisher diagnostics are provided in Appendix~\ref{app:geometry_details}.

\textbf{Gradient alignment.} We compare average log-likelihood gradients of each anchor's own correct math responses and shared corrupted trajectories using layer-wise cosines. A large positive cosine means the interference data pushes along the direction that would further improve capability.

\section{Experiments}
\label{sec:experiments}

\begin{table*}[!t]
\centering
\caption{Domain-dependent interference results. Accuracy (\%, $\uparrow$) and macro retention relative to each anchor (\%, $\uparrow$). Shared no-interference baselines are listed once, followed by the three interference conditions. HE+ and MBPP+ report code performance. Dashes indicate values not reported.}
\label{tab:main_results}
\footnotesize
\setlength{\tabcolsep}{3pt}
\renewcommand{\arraystretch}{1.08}
\begin{tabularx}{\textwidth}{@{}>{\columncolor{white}[0pt][\tabcolsep]}ll*{5}{>{\centering\arraybackslash}X}>{\columncolor{white}[\tabcolsep][0pt]\centering\arraybackslash}X@{}}
\toprule
 & & \multicolumn{4}{c}{Math} & \multicolumn{2}{c}{Code} \\
\cmidrule(lr){3-6}\cmidrule(lr){7-8}
Anchor & Defense & MATH500 & AIME24 & AIME25 & MacroRet & HE+ & MBPP+ \\
\specialrule{\lightrulewidth}{0pt}{0pt}
\rowcolor{tablebaseline}
\multicolumn{8}{@{}c@{}}{\textit{No interference (shared baselines)}} \\
- & Base & 73.00 & 10.83 & 6.25 & - & 69.5 & 65.6 \\
\rowcolor{tablerowgray}
\GRPO & Anchor & 85.40 & 24.17 & 21.67 & 100.00 & 75.0 & 65.3 \\
SFT & Anchor & 81.55 & 23.96 & 25.42 & 100.00 & 78.7 & 54.8 \\
\specialrule{\lightrulewidth}{0pt}{0pt}
\rowcolor{tablematched}
\multicolumn{8}{@{}c@{}}{\textit{Matched domain: incorrect math trajectories}} \\
\GRPO & None & 56.60 & 7.92 & 10.83 & 49.68 & 72.0 & 64.8 \\
\rowcolor{tablerowgray}
SFT & None & 61.15 & 8.33 & 10.21 & 49.98 & 72.6 & 60.3 \\
\GRPO & EWC $10^{6}$ & 66.30 & 13.33 & 11.25 & 61.57 & \textbf{75.0} & \underline{65.1} \\
\rowcolor{tablerowgray}
SFT & EWC $10^{6}$ & 63.15 & 11.88 & 12.08 & 58.18 & 73.8 & 58.2 \\
\GRPO & EWC $10^{7}$ & \textbf{76.60} & \underline{15.63} & \underline{15.42} & \underline{75.17} & \underline{74.4} & \textbf{66.4} \\
\rowcolor{tablerowgray}
SFT & EWC $10^{7}$ & \underline{74.45} & \textbf{17.50} & \textbf{16.88} & \textbf{76.91} & 73.8 & 56.9 \\
\specialrule{\lightrulewidth}{0pt}{0pt}
\rowcolor{tablenear}
\multicolumn{8}{@{}c@{}}{\textit{Near domain: code trajectories}} \\
\GRPO & None & \underline{82.35} & 18.75 & 17.08 & \underline{84.29} & 54.3 & 59.3 \\
\rowcolor{tablerowgray}
SFT & None & 73.60 & 18.75 & 16.67 & 78.03 & \underline{56.7} & \underline{59.5} \\
\GRPO & EWC $10^{6}$ & \textbf{84.00} & \textbf{20.83} & \textbf{21.25} & \textbf{94.20} & \textbf{59.8} & \textbf{59.8} \\
\rowcolor{tablerowgray}
SFT & EWC $10^{6}$ & 75.40 & \underline{19.58} & \underline{17.71} & 81.28 & 55.5 & 58.7 \\
\specialrule{\lightrulewidth}{0pt}{0pt}
\rowcolor{tablefar}
\multicolumn{8}{@{}c@{}}{\textit{Far domain: XSum summarization}} \\
\GRPO & None & \textbf{84.95} & \textbf{24.17} & \textbf{18.54} & \textbf{95.02} & \textbf{75.6} & \textbf{67.5} \\
\rowcolor{tablerowgray}
SFT & None & \underline{79.10} & \underline{14.79} & \underline{15.00} & \underline{72.58} & \underline{75.0} & \underline{62.2} \\
\bottomrule
\end{tabularx}
\end{table*}

We evaluate whether the post-training objective used to instill a capability dictates its resilience against subsequent learning. Accordingly, our experimental protocol comprises two stages. In the anchoring stage, we train two mathematical reasoning anchors from the same base model—one via supervised fine-tuning (SFT) and the other via Group Relative Policy Optimization (GRPO). In the interference stage, we continue training each anchor on data deliberately chosen to conflict with the anchored capability, quantifying the subsequent retention of that capability. Interference data are sampled at three distinct domain distances from the anchor domain, enabling us to determine whether any observed stability differences are global or domain-selective. We first describe the experimental setup, then present the primary empirical findings and analyze the underlying structural mechanisms.

\subsection{Experimental Settings}
\label{sec:settings}

\textbf{Models and Data.} We use Qwen3-4B~\citep{Yang2025Qwen3TR} and SmolLM3-3B~\citep{bakouch2025smollm3} as base models. All experimental results presented in the main text are obtained with Qwen3-4B. Results for SmolLM3-3B are reported in Appendix~\ref{app:experiment_scale}, Table~\ref{tab:smollm}. Additional results for Qwen3-8B-Base are reported in Appendix~\ref{app:qwen3_8b}. Both SFT and \GRPO\ anchors are trained on the first 19,000 examples of Math-CoT-20k~\citep{Ren2026MathCoT}, with reference trajectories from Qwen3-32B. A holdout set of 1,480 examples is reserved, including 1,024 prompts for Fisher estimation. Anchor performance is reported in Table~\ref{tab:main_results}. Interference data are organized by domain distance from the anchored math task. The \textit{matched} condition uses incorrect math trajectories from OpenR1-Math-220k, the \textit{near} condition uses code-error trajectories from APPS (plus controlled trajectories based on APPS+~\citep{Dou2024StepCoder}), and the \textit{far} condition uses XSum data.

\textbf{Evaluation.} We evaluate mathematical retention on MATH500~\citep{Hendrycks2021MATH,Lightman2023Verify}, AIME24, and AIME25~\citep{MAAAIME}, and additionally report HumanEval+ and MBPP+~\citep{Chen2021HumanEval,Austin2021MBPP,Liu2023EvalPlus} under code interference. Retention is measured with $\mathrm{MacroRet}$ as defined in Equation~\ref{eq:ret}. Parameter-space ablations additionally use $\mathrm{GainPres}$ as defined in Equation~\ref{eq:gainpres}. We also track the macro forward KL divergence to the base policy on 1,000 fixed contexts per probe domain.

\textbf{Conditions and Defenses.} Each anchor serves as its own no-interference upper bound. During interference training, we compare vanilla continued SFT, diagonal EWC with $\lambda\in\{10^6,10^7\}$, and freezing transformer layers 32-34. EWC uses the Fisher diagonal estimated from each anchor's held-out math prompts.

\subsection{Main Results}
\label{sec:main_results}

\paragraph{Matched-domain interference.}

The matched-domain block of Table~\ref{tab:main_results} reports continued training on incorrect math trajectories, the condition closest to a same-domain attack on the anchored capability. Retention is severe for both anchors: roughly half of the anchored math accuracy is lost without a regularizer. The two objectives have similar point estimates: at $\lambda=0$ the macro retention is 49.68\% for \GRPO\ against 49.98\% for SFT. As the penalty is strengthened, retention improves monotonically for both anchors, but the \GRPO\ advantage does not emerge. At $\lambda=10^{7}$, SFT is marginally ahead (76.91\% versus 75.17\%). The no-defense GRPO SFT gap is $-0.30$ percentage points.

\paragraph{Near-domain interference.}

The near-domain block of Table~\ref{tab:main_results} reports interference from code trajectories. Without any defense, the \GRPO\ anchor retains 84.29\% of its math ability against 78.03\% for the SFT anchor, a gap of 6.26 percentage points. Adding EWC at $\lambda=10^{6}$ lifts both anchors, and it lifts the \GRPO\ anchor more: retention reaches 94.20\% against 81.28\%, widening the gap to 12.92 percentage points. The collateral code columns show a different pattern: neither interference nor defense produces a consistent change on HumanEval+ or MBPP+, indicating that the effect we measure is specific to the anchored math capability rather than to general degradation.

\paragraph{Far-domain interference.}

Under continued summarization training, the results in Table~\ref{tab:main_results} show that the \GRPO\ anchor loses almost nothing: AIME24 is unchanged at 24.17\% and MATH500 falls by less than half a point (85.40\% to 84.95\%), whereas the SFT anchor loses roughly 27\% of its relative math accuracy (72.58\% retention). The GRPO SFT gap is 22.43 percentage points. Details of domain interaction and controlled corruption are provided in Appendices~\ref{app:paired_bootstrap} and~\ref{app:experiment_corruption}.

Appendix~\ref{app:on_policy_sft} reports an additional on-policy SFT control trained on correct GRPO-generated trajectories that recovers much of GRPO's cross-domain retention, but exhibits substantially less stable generation under matched-domain interference, including frequent excessively long responses.

\subsection{Analysis}
\label{sec:analysis}

\begin{wraptable}{R}{0.5\textwidth}
\vspace{-\intextsep}
\centering
\caption{Spectral ablation of the anchor update at matched budgets. Bold and underline mark the largest and second-largest entry of the last two columns.}
\label{tab:spectral}

\footnotesize
\setlength{\tabcolsep}{1.5pt}
\renewcommand{\arraystretch}{1.05}
\begin{tabularx}{\linewidth}{@{}>{\columncolor{white}[0pt][\tabcolsep]}ll*{3}{>{\centering\arraybackslash}X}>{\columncolor{white}[\tabcolsep][0pt]\centering\arraybackslash}X@{}}
\toprule
 & & \multicolumn{2}{c}{\shortstack{Energy\\carried (\%)}} & \multicolumn{2}{c}{GainPres (\%)} \\
\cmidrule(lr){3-4}
\cmidrule(lr){5-6}
Anchor & Budget & Keep & Remove & \shortstack{Keep\\($\uparrow$)} & \shortstack{Remove\\($\uparrow$)} \\
\midrule
\GRPO & Top-1 & 6.72 & 6.72 & 31.01 & \underline{80.98} \\
\rowcolor{tablerowgray}
SFT & Top-1 & 1.06 & 1.06 & $-0.23$ & \textbf{94.57} \\
\GRPO & Top-8 & 18.50 & 18.50 & 83.10 & 70.25 \\
\rowcolor{tablerowgray}
SFT & Top-8 & 3.47 & 3.47 & 27.35 & 77.03 \\
\GRPO & Top-32 & 32.30 & 32.30 & 89.46 & 54.80 \\
\rowcolor{tablerowgray}
SFT & Top-32 & 8.11 & 8.11 & 45.52 & 78.82 \\
\GRPO & Energy 25\% & 25.37 & 25.37 & 87.61 & 42.70 \\
\rowcolor{tablerowgray}
SFT & Energy 25\% & 25.05 & 25.05 & \underline{105.85} & 59.00 \\
\GRPO & Energy 50\% & 50.09 & 50.09 & 98.28 & 25.54 \\
\rowcolor{tablerowgray}
SFT & Energy 50\% & 50.03 & 50.03 & \textbf{111.36} & 50.82 \\
\GRPO & Energy 75\% & 75.02 & 75.02 & 95.41 & 21.71 \\
\rowcolor{tablerowgray}
SFT & Energy 75\% & 75.02 & 75.02 & 101.31 & $-3.95$ \\
\bottomrule
\end{tabularx}
\end{wraptable}

\paragraph{Ablation: where the gain lives.}

Two measurements reported in Table~\ref{tab:geometry} frame the mechanism. The two updates reach comparable math performance from the same base model yet are very nearly orthogonal in parameter space, so they do not implement the same solution. The \GRPO\ update is also markedly more concentrated, requiring only about 59\% of the directions that the SFT update needs to carry 90\% of its own energy. Table~\ref{tab:spectral} decomposes the anchor update $\Delta W=U\Sigma V^{\top}$ and tests whether the leading directions are sufficient (keep) or necessary (remove) for the post-training gain. ``Energy'' is the fraction of squared singular-value mass captured by the kept directions (keep) or removed by the deleted directions (remove), and GainPres, defined in Equation~\eqref{eq:gainpres}, is the fraction of the anchor gain recovered under keep and the fraction retained under remove. Values above 100\% indicate that the perturbed model \emph{exceeds} its own parent in score. This does not imply an absolute improvement across anchors, so comparison within a budget row is the intended reading. The two anchors give qualitatively different answers. At a fixed rank budget, keeping the top-8 singular directions of each analyzed update matrix preserves 83.10\% of the \GRPO\ gain versus 27.35\% of the SFT gain, as visualized in the inset of Figure~\ref{fig:retention}(a). Keeping half of the \GRPO\ update's energy already reproduces nearly all of its gain (98.28\% of the base-to-anchor gain), whereas the same budget applied to SFT yields a perturbed model that overshoots its own parent (111.36\%). Deleting the leading singular direction of each analyzed update matrix removes more of \GRPO's gain than of SFT's (GainPres 80.98\% versus 94.57\%). This ordering persists at 50\% energy removed (\GRPO\ 25.54\% versus SFT 50.82\%) and reverses at 75\% removed (\GRPO\ 21.71\% versus SFT $-3.95\%$). Together these observations say that the \GRPO\ update is both highly compressible and highly concentrated: most of the capability can be written in few directions, and exactly for that reason, it depends on those directions. Table~\ref{tab:geometry} quantifies the same difference statically. Table~\ref{tab:apps} and the far-domain results in Table~\ref{tab:main_results} show that this concentration manifests as retention when the attack comes from outside the anchor's domain. Appendix~\ref{app:experiment_fisher} provides high-Fisher-coordinate analysis.

\begin{wrapfigure}{R}{0.5\textwidth}
\vspace{-\intextsep}
\centering
\includegraphics[width=\linewidth]{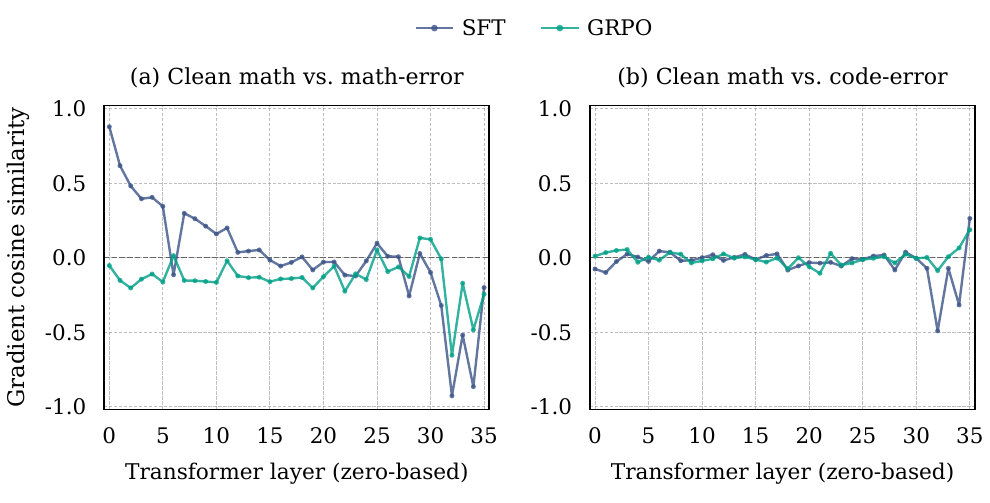}
\caption{Layer-wise cosine similarity between the gradient of each anchor's own correct responses and the gradient of shared corrupted trajectories, for the math and the code corruption set.}
\label{fig:gradsim}
\end{wrapfigure}

\paragraph{Why the difference, and why it is domain-dependent.}

Figure~\ref{fig:gradsim} adds the dynamic counterpart. For SFT, the gradient of clean math data and the gradient of incorrect math trajectories are strongly aligned in the early layers, reaching roughly 0.87 at the first layer and staying in the 0.2-0.5 range over the first ten layers. For \GRPO, the alignment is near zero almost everywhere, with the exception of the last few layers where both oscillate sharply. A continued SFT step on wrong trajectories therefore pushes along a direction that the clean and the corrupted gradients share, whereas the \GRPO\ update is largely indifferent to it. This explains the near- and far-domain separation in Table~\ref{tab:main_results}. It also explains why the separation vanishes in the matched-domain block of Table~\ref{tab:main_results}: when the corruption is itself drawn from the anchor domain, the interference gradient can reach the capability even for \GRPO, which is exactly the regime in which Figure~\ref{fig:kl} shows \GRPO\ drifting more than SFT. We therefore attribute the advantage to interference acting in a direction that the objective kept largely unused, rather than to any absolute robustness of \GRPO-derived capabilities.

\paragraph{Distributional drift is gated by domain match.}
Figure~\ref{fig:kl} traces the macro forward KL divergence to the base policy across ten interference steps, evaluated on fixed contexts from each of the three probe domains. On mismatched probes the picture is consistent with prior accounts of RL as a conservative update: the \GRPO\ anchor stays near the base model throughout (approximately 0.02-0.05) while the SFT anchor drifts and only slowly recovers. On the two matched probes, however, this ordering reverses: with code interference probed on code contexts, \GRPO\ rises from about 0.02 to about 0.14 while SFT moves from about 0.13 down to about 0.08, and with summarization interference probed on summarization contexts \GRPO\ rises from about 0.045 to about 0.31 while SFT changes from about 0.36 to about 0.40. Drift is therefore not a monotone function of the post-training objective. It depends on whether the interference acts within the domain that carries the anchored capability. Further observations are provided in Appendix~\ref{app:experiment_observations}.

\section{Conclusion}
We asked whether the post-training objective that produces a capability determines how well that capability survives subsequent learning. Our answer is deliberately narrower than the prevailing narrative: it does, but only as a function of what the model is trained on next. Under matched-domain interference, token-matched SFT and GRPO math anchors are statistically inseparable. Under far-domain interference, they differ by more than twenty points of normalized retention, and the far-minus-matched interaction is large and positive. The advantage of RL over SFT, documented by a growing literature, is therefore domain-selective rather than global. Our parameter-space analysis suggests why: the two objectives reach comparable math performance via nearly orthogonal, differently concentrated updates, and the concentrated GRPO solution is simultaneously more compressible and more dependent on a small set of dominant directions. In practice, the implication is that the choice of objective cannot by itself ensure a deployed capability against future training. What matters is the relationship between the capability to be protected and the data the model will consume next. \textit{\textbf{RETENTION PROBE}} is our concrete contribution toward making this relationship measurable: a protocol combining graded interference domains, anchor-relative retention, and interaction-based inference. Future work should extend the probe to more anchor domains, additional model families and scales, and interference delivered by on-policy objectives, and should test whether the geometry we characterize can be enforced as a constraint rather than merely observed.

\section*{Limitations}
\label{sec:limitations}
The two anchors differ in on-policy versus offline targets and optimization
settings, so the comparison does not isolate the training objective alone.
Results for the additional on-policy SFT control are reported in
Appendix~\ref{app:on_policy_sft}.
The prompt-domain analysis uses one frozen encoder. Its distances may reflect
topic, residual format, and input length as well as task relation. Bandwidth
and truncation checks do not establish encoder-independent robustness. Its
bootstrap ranges condition on the encoder and selected bandwidth, and the raw
percentile procedure is biased near zero.

\subsection*{AI use statement}
AI-assisted tools were used to polish the language and improve the clarity of
the manuscript. The authors take full responsibility for the scientific
content, the accuracy of the reported results, and the final text.

\subsection*{Reproducibility statement}
Our study supports a high degree of reproducibility through detailed
documentation of the experimental setup, evaluation procedures, and analysis
methods, enabling researchers to reconstruct the experiments and verify the
reported findings.
We explicitly define the experimental protocol and evaluation
metrics. Appendix~\ref{app:anchor_details} documents the anchor data splits,
training configurations, checkpoint selection, and token accounting, while
Appendix~\ref{app:implementation} provides interference settings, evaluation
procedures, software details, and runtime information. The paired-bootstrap
procedure is specified in Appendix~\ref{app:paired_bootstrap}, and the
parameter-space diagnostics and prompt-domain distance analysis are detailed
in Appendices~\ref{app:geometry_details} and~\ref{app:domain_distance}.

\subsection*{Ethics statement}
This work studies capability retention in language models through controlled
computational experiments on mathematical reasoning, code, and summarization
tasks. Understanding when subsequent training degrades existing capabilities
can inform more reliable model adaptation. However, the same diagnostics
could also be used to identify and deliberately disrupt model capabilities.
The incorrect mathematical and code trajectories used in our interference
experiments are research stress tests and should not be treated as verified
solutions or reliable training targets. Our evaluations concern task
performance and do not establish the preservation of safety alignment,
fairness, or privacy during continued training. These properties require
separate assessment before deployment.

\begingroup
\setlength{\bibsep}{3pt plus 0.3pt minus 0.3pt}
\bibliography{iclr2027_conference}

\begin{thebibliography}{49}
\providecommand{\natexlab}[1]{#1}
\providecommand{\url}[1]{\texttt{#1}}
\expandafter\ifx\csname urlstyle\endcsname\relax
  \providecommand{\doi}[1]{doi: #1}\else
  \providecommand{\doi}{doi: \begingroup \urlstyle{rm}\Url}\fi

\bibitem[Austin et~al.(2021)Austin, Odena, Nye, Bosma, Michalewski, Dohan, Jiang, Cai, Terry, Le, and Sutton]{Austin2021MBPP}
Jacob Austin, Augustus Odena, Maxwell Nye, Maarten Bosma, Henryk Michalewski, David Dohan, Ellen Jiang, Carrie Cai, Michael Terry, Quoc Le, and Charles Sutton.
\newblock Program synthesis with large language models.
\newblock \emph{arXiv preprint arXiv:2108.07732}, 2021.
\newblock URL \url{https://arxiv.org/abs/2108.07732}.

\bibitem[Bakouch et~al.(2025)Bakouch, Ben~Allal, Lozhkov, Tazi, Tunstall, Patiño, Beeching, Roucher, Reedi, Gallouédec, Rasul, Habib, Fourrier, Kydlicek, Penedo, Larcher, Morlon, Srivastav, Lochner, Nguyen, Raffel, von Werra, and Wolf]{bakouch2025smollm3}
Elie Bakouch, Loubna Ben~Allal, Anton Lozhkov, Nouamane Tazi, Lewis Tunstall, Carlos~Miguel Patiño, Edward Beeching, Aymeric Roucher, Aksel~Joonas Reedi, Quentin Gallouédec, Kashif Rasul, Nathan Habib, Clémentine Fourrier, Hynek Kydlicek, Guilherme Penedo, Hugo Larcher, Mathieu Morlon, Vaibhav Srivastav, Joshua Lochner, Xuan-Son Nguyen, Colin Raffel, Leandro von Werra, and Thomas Wolf.
\newblock {SmolLM3: smol, multilingual, long-context reasoner}.
\newblock \url{https://huggingface.co/blog/smollm3}, 2025.

\bibitem[Biderman et~al.(2024)Biderman, Portes, Ortiz, Paul, Greengard, Jennings, King, Havens, Chiley, Frankle, Blakeney, and Cunningham]{Biderman2024LoRALL}
Dan Biderman, Jacob Portes, Jose Javier~Gonzalez Ortiz, Mansheej Paul, Philip Greengard, Connor Jennings, Daniel King, Sam Havens, Vitaliy Chiley, Jonathan Frankle, Cody Blakeney, and John~P. Cunningham.
\newblock {LoRA} learns less and forgets less.
\newblock \emph{Transactions on Machine Learning Research}, 2024.
\newblock URL \url{https://arxiv.org/abs/2405.09673}.

\bibitem[Chen et~al.(2026)Chen, Razin, Narasimhan, and Chen]{Chen2025RetainingBD}
Howard Chen, Noam Razin, Karthik Narasimhan, and Danqi Chen.
\newblock Retaining by doing: The role of on-policy data in mitigating forgetting.
\newblock In \emph{Proceedings of the 43rd International Conference on Machine Learning}, 2026.
\newblock URL \url{https://arxiv.org/abs/2510.18874}.

\bibitem[Chen et~al.(2021)Chen, Tworek, Jun, Yuan, de~Oliveira~Pinto, Kaplan, Edwards, Burda, Joseph, Brockman, Ray, Puri, Krueger, Petrov, Khlaaf, Sastry, Mishkin, Chan, Gray, Ryder, Pavlov, Power, Kaiser, Bavarian, Winter, Tillet, Such, Cummings, Plappert, Chantzis, Barnes, Herbert-Voss, Guss, Nichol, Paino, Tezak, Tang, Babuschkin, Balaji, Jain, Saunders, Hesse, Carr, Leike, Achiam, Misra, Morikawa, Radford, Knight, Brundage, Murati, Mayer, Welinder, McGrew, Amodei, McCandlish, Sutskever, and Zaremba]{Chen2021HumanEval}
Mark Chen, Jerry Tworek, Heewoo Jun, Qiming Yuan, Henrique~Ponde de~Oliveira~Pinto, Jared Kaplan, Harri Edwards, Yuri Burda, Nicholas Joseph, Greg Brockman, Alex Ray, Raul Puri, Gretchen Krueger, Michael Petrov, Heidy Khlaaf, Girish Sastry, Pamela Mishkin, Brooke Chan, Scott Gray, Nick Ryder, Mikhail Pavlov, Alethea Power, Lukasz Kaiser, Mohammad Bavarian, Clemens Winter, Philippe Tillet, Felipe~Petroski Such, Dave Cummings, Matthias Plappert, Fotios Chantzis, Elizabeth Barnes, Ariel Herbert-Voss, William~Hebgen Guss, Alex Nichol, Alex Paino, Nikolas Tezak, Jie Tang, Igor Babuschkin, Suchir Balaji, Shantanu Jain, William Saunders, Christopher Hesse, Andrew~N. Carr, Jan Leike, Josh Achiam, Vedant Misra, Evan Morikawa, Alec Radford, Matthew Knight, Miles Brundage, Mira Murati, Katie Mayer, Peter Welinder, Bob McGrew, Dario Amodei, Sam McCandlish, Ilya Sutskever, and Wojciech Zaremba.
\newblock Evaluating large language models trained on code.
\newblock \emph{arXiv preprint arXiv:2107.03374}, 2021.
\newblock URL \url{https://arxiv.org/abs/2107.03374}.

\bibitem[Cheng et~al.(2025)Cheng, Hao, Liu, Zhou, Xie, Yao, Bian, Zhuang, Dey, Zha, Gu, Zhou, Wang, Li, Fan, She, Gao, Saparov, Li, Killian, Yurochkin, Liu, Xing, and Hu]{Cheng2025RevisitingRL}
Zhoujun Cheng, Shibo Hao, Tianyang Liu, Fan Zhou, Yutao Xie, Feng Yao, Yuexin Bian, Yonghao Zhuang, Nilabjo Dey, Yuheng Zha, Yi~Gu, Kun Zhou, Yuqi Wang, Yuan Li, Richard Fan, Jianshu She, Chengqian Gao, Abulhair Saparov, Haonan Li, Taylor~W. Killian, Mikhail Yurochkin, Zhengzhong Liu, Eric~P. Xing, and Zhiting Hu.
\newblock Revisiting reinforcement learning for {LLM} reasoning from a cross-domain perspective.
\newblock \emph{arXiv preprint arXiv:2506.14965}, 2025.
\newblock URL \url{https://arxiv.org/abs/2506.14965}.

\bibitem[Chu et~al.(2025)Chu, Zhai, Yang, Tong, Xie, Schuurmans, Le, Levine, and Ma]{Chu2025SFTMR}
Tianzhe Chu, Yuexiang Zhai, Jihan Yang, Shengbang Tong, Saining Xie, Dale Schuurmans, Quoc~V. Le, Sergey Levine, and Yi~Ma.
\newblock {SFT} memorizes, {RL} generalizes: A comparative study of foundation model post-training.
\newblock \emph{arXiv preprint arXiv:2501.17161}, 2025.
\newblock URL \url{https://arxiv.org/abs/2501.17161}.

\bibitem[{DeepSeek-AI}(2025)]{DeepSeekAI2025DeepSeekR1IR}
{DeepSeek-AI}.
\newblock {DeepSeek-R1}: Incentivizing reasoning capability in {LLMs} via reinforcement learning.
\newblock \emph{arXiv preprint arXiv:2501.12948}, 2025.
\newblock URL \url{https://arxiv.org/abs/2501.12948}.

\bibitem[{DeepSeek-AI}(2026)]{DeepSeek2026V4}
{DeepSeek-AI}.
\newblock {DeepSeek-V4}: Towards highly efficient million-token context intelligence.
\newblock \emph{arXiv preprint arXiv:2606.19348}, 2026.
\newblock URL \url{https://arxiv.org/abs/2606.19348}.

\bibitem[Dou et~al.(2024)Dou, Liu, Jia, Zhou, Xiong, Shan, Huang, Wang, Fan, Xi, Zhou, Ji, Zheng, Zhang, Gui, and Huang]{Dou2024StepCoder}
Shihan Dou, Yan Liu, Haoxiang Jia, Enyu Zhou, Limao Xiong, Junjie Shan, Caishuang Huang, Xiao Wang, Xiaoran Fan, Zhiheng Xi, Yuhao Zhou, Tao Ji, Rui Zheng, Qi~Zhang, Tao Gui, and Xuanjing Huang.
\newblock {S}tep{C}oder: Improving code generation with reinforcement learning from compiler feedback.
\newblock In Lun-Wei Ku, Andre Martins, and Vivek Srikumar, editors, \emph{Proceedings of the 62nd Annual Meeting of the Association for Computational Linguistics (Volume 1: Long Papers)}, pages 4571--4585, Bangkok, Thailand, August 2024. Association for Computational Linguistics.
\newblock \doi{10.18653/v1/2024.acl-long.251}.
\newblock URL \url{https://aclanthology.org/2024.acl-long.251/}.

\bibitem[Feng et~al.(2025)Feng, Constable, and Mao]{PyTorchFSDP2}
Wei Feng, Will Constable, and Yifan Mao.
\newblock Getting started with fully sharded data parallel ({FSDP2}).
\newblock PyTorch Tutorials, 2025.
\newblock URL \url{https://docs.pytorch.org/tutorials/intermediate/FSDP_tutorial.html}.
\newblock Accessed September 23, 2026.

\bibitem[Gretton et~al.(2012)Gretton, Borgwardt, Rasch, Sch{\"o}lkopf, and Smola]{Gretton2012MMD}
Arthur Gretton, Karsten~M. Borgwardt, Malte~J. Rasch, Bernhard Sch{\"o}lkopf, and Alexander Smola.
\newblock A kernel two-sample test.
\newblock \emph{Journal of Machine Learning Research}, 13\penalty0 (25):\penalty0 723--773, 2012.
\newblock URL \url{https://jmlr.org/papers/v13/gretton12a.html}.

\bibitem[Hendrycks et~al.(2021{\natexlab{a}})Hendrycks, Basart, Kadavath, Mazeika, Arora, Guo, Burns, Puranik, He, Song, and Steinhardt]{Hendrycks2021APPS}
Dan Hendrycks, Steven Basart, Saurav Kadavath, Mantas Mazeika, Akul Arora, Ethan Guo, Collin Burns, Samir Puranik, Horace He, Dawn Song, and Jacob Steinhardt.
\newblock Measuring coding challenge competence with {APPS}.
\newblock \emph{arXiv preprint arXiv:2105.09938}, 2021{\natexlab{a}}.
\newblock URL \url{https://arxiv.org/abs/2105.09938}.

\bibitem[Hendrycks et~al.(2021{\natexlab{b}})Hendrycks, Burns, Kadavath, Arora, Basart, Tang, Song, and Steinhardt]{Hendrycks2021MATH}
Dan Hendrycks, Collin Burns, Saurav Kadavath, Akul Arora, Steven Basart, Eric Tang, Dawn Song, and Jacob Steinhardt.
\newblock Measuring mathematical problem solving with the {MATH} dataset.
\newblock \emph{arXiv preprint arXiv:2103.03874}, 2021{\natexlab{b}}.
\newblock URL \url{https://arxiv.org/abs/2103.03874}.

\bibitem[Ilharco et~al.(2022)Ilharco, Ribeiro, Wortsman, Gururangan, Schmidt, Hajishirzi, and Farhadi]{Ilharco2022EditingMW}
Gabriel Ilharco, Marco~Tulio Ribeiro, Mitchell Wortsman, Suchin Gururangan, Ludwig Schmidt, Hannaneh Hajishirzi, and Ali Farhadi.
\newblock Editing models with task arithmetic.
\newblock \emph{arXiv preprint arXiv:2212.04089}, 2022.
\newblock URL \url{https://arxiv.org/abs/2212.04089}.

\bibitem[Kirkpatrick et~al.(2017)Kirkpatrick, Pascanu, Rabinowitz, Veness, Desjardins, Rusu, Milan, Quan, Ramalho, Grabska-Barwinska, Hassabis, Clopath, Kumaran, and Hadsell]{Kirkpatrick2016OvercomingCF}
James Kirkpatrick, Razvan Pascanu, Neil~C. Rabinowitz, Joel Veness, Guillaume Desjardins, Andrei~A. Rusu, Kieran Milan, John Quan, Tiago Ramalho, Agnieszka Grabska-Barwinska, Demis Hassabis, Claudia Clopath, Dharshan Kumaran, and Raia Hadsell.
\newblock Overcoming catastrophic forgetting in neural networks.
\newblock \emph{Proceedings of the National Academy of Sciences}, 114\penalty0 (13):\penalty0 3521--3526, 2017.
\newblock \doi{10.1073/pnas.1611835114}.
\newblock URL \url{https://doi.org/10.1073/pnas.1611835114}.

\bibitem[Kwon et~al.(2023)Kwon, Li, Zhuang, Sheng, Zheng, Yu, Gonzalez, Zhang, and Stoica]{Kwon2023vLLM}
Woosuk Kwon, Zhuohan Li, Siyuan Zhuang, Ying Sheng, Lianmin Zheng, Cody~Hao Yu, Joseph~E. Gonzalez, Hao Zhang, and Ion Stoica.
\newblock Efficient memory management for large language model serving with {PagedAttention}.
\newblock \emph{arXiv preprint arXiv:2309.06180}, 2023.
\newblock URL \url{https://arxiv.org/abs/2309.06180}.

\bibitem[Kydl{\'i}{\v c}ek(2025)]{Kydlicek2025MathVerify}
Hynek Kydl{\'i}{\v c}ek.
\newblock {Math-Verify}: Math verification library.
\newblock GitHub repository, 2025.
\newblock URL \url{https://github.com/huggingface/Math-Verify}.

\bibitem[Lee et~al.(2021)Lee, Goldt, and Saxe]{Lee2021ContinualLearning}
Sebastian Lee, Sebastian Goldt, and Andrew Saxe.
\newblock Continual learning in the teacher-student setup: Impact of task similarity.
\newblock In \emph{Proceedings of the 38th International Conference on Machine Learning}, volume 139 of \emph{Proceedings of Machine Learning Research}, pages 6109--6119. PMLR, 2021.
\newblock URL \url{https://proceedings.mlr.press/v139/lee21e.html}.

\bibitem[Li and Hoiem(2018)]{Li2016LearningWF}
Zhizhong Li and Derek Hoiem.
\newblock Learning without forgetting.
\newblock \emph{IEEE Transactions on Pattern Analysis and Machine Intelligence}, 40\penalty0 (12):\penalty0 2935--2947, 2018.
\newblock \doi{10.1109/TPAMI.2017.2773081}.
\newblock URL \url{https://doi.org/10.1109/TPAMI.2017.2773081}.

\bibitem[Lightman et~al.(2023)Lightman, Kosaraju, Burda, Edwards, Baker, Lee, Leike, Schulman, Sutskever, and Cobbe]{Lightman2023Verify}
Hunter Lightman, Vineet Kosaraju, Yura Burda, Harri Edwards, Bowen Baker, Teddy Lee, Jan Leike, John Schulman, Ilya Sutskever, and Karl Cobbe.
\newblock Let's verify step by step.
\newblock \emph{arXiv preprint arXiv:2305.20050}, 2023.
\newblock URL \url{https://arxiv.org/abs/2305.20050}.

\bibitem[Liu et~al.(2023)Liu, Xia, Wang, and Zhang]{Liu2023EvalPlus}
Jiawei Liu, Chunqiu~Steven Xia, Yuyao Wang, and Lingming Zhang.
\newblock Is your code generated by {ChatGPT} really correct? rigorous evaluation of large language models for code generation.
\newblock In \emph{Advances in Neural Information Processing Systems}, 2023.
\newblock URL \url{https://arxiv.org/abs/2305.01210}.

\bibitem[Liu et~al.(2026)Liu, Yin, Wang, Fan, Shi, Li, Gao, and Tao]{Liu2026UnderstandingAE}
Shangge Liu, Yuehan Yin, Lei Wang, Qi~Fan, Yinghuan Shi, Wenbin Li, Yang Gao, and Dacheng Tao.
\newblock Understanding and enforcing weight disentanglement in task arithmetic.
\newblock \emph{arXiv preprint arXiv:2604.17078}, 2026.
\newblock URL \url{https://arxiv.org/abs/2604.17078}.

\bibitem[Loshchilov and Hutter(2019)]{Loshchilov2019AdamW}
Ilya Loshchilov and Frank Hutter.
\newblock Decoupled weight decay regularization.
\newblock In \emph{International Conference on Learning Representations}, 2019.
\newblock URL \url{https://arxiv.org/abs/1711.05101}.

\bibitem[Luo et~al.(2025)Luo, Yang, Meng, Li, Zhou, and Zhang]{Luo2023AnES}
Yun Luo, Zhen Yang, Fandong Meng, Yafu Li, Jie Zhou, and Yue Zhang.
\newblock An empirical study of catastrophic forgetting in large language models during continual fine-tuning.
\newblock \emph{IEEE Transactions on Audio, Speech and Language Processing}, 33:\penalty0 3776--3786, 2025.
\newblock \doi{10.1109/TASLPRO.2025.3606231}.
\newblock URL \url{https://doi.org/10.1109/TASLPRO.2025.3606231}.

\bibitem[{Mathematical Association of America}(n.d.)]{MAAAIME}
{Mathematical Association of America}.
\newblock American invitational mathematics examination ({AIME}).
\newblock Official competition website, n.d.
\newblock URL \url{https://maa.org/maa-invitational-competitions/}.
\newblock 2024 and 2025 examinations; accessed September 23, 2026.

\bibitem[McCloskey and Cohen(1989)]{McCloskey1989CatastrophicII}
Michael McCloskey and Neal~J. Cohen.
\newblock Catastrophic interference in connectionist networks: The sequential learning problem.
\newblock In \emph{Psychology of Learning and Motivation}, volume~24, pages 109--165. Academic Press, 1989.
\newblock \doi{10.1016/S0079-7421(08)60536-8}.
\newblock URL \url{https://doi.org/10.1016/S0079-7421(08)60536-8}.

\bibitem[Mukherjee et~al.(2025)Mukherjee, Yuan, Hakkani-Tur, and Peng]{Mukherjee2025ReinforcementLF}
Sagnik Mukherjee, Lifan Yuan, Dilek Hakkani-Tur, and Hao Peng.
\newblock Reinforcement learning finetunes small subnetworks in large language models.
\newblock In \emph{Advances in Neural Information Processing Systems}, volume~38, 2025.
\newblock URL \url{https://arxiv.org/abs/2505.11711}.

\bibitem[Narayan et~al.(2018)Narayan, Cohen, and Lapata]{Narayan2018XSum}
Shashi Narayan, Shay~B. Cohen, and Mirella Lapata.
\newblock Don{'}t give me the details, just the summary! topic-aware convolutional neural networks for extreme summarization.
\newblock In Ellen Riloff, David Chiang, Julia Hockenmaier, and Jun{'}ichi Tsujii, editors, \emph{Proceedings of the 2018 Conference on Empirical Methods in Natural Language Processing}, pages 1797--1807, Brussels, Belgium, October-November 2018. Association for Computational Linguistics.
\newblock \doi{10.18653/v1/D18-1206}.
\newblock URL \url{https://aclanthology.org/D18-1206/}.

\bibitem[{Open R1 Team}(2025)]{OpenR1Math2025}
{Open R1 Team}.
\newblock {OpenR1-Math-220k}.
\newblock Hugging Face dataset, 2025.
\newblock URL \url{https://huggingface.co/datasets/open-r1/OpenR1-Math-220k}.

\bibitem[Ouyang et~al.(2022)Ouyang, Wu, Jiang, Almeida, Wainwright, Mishkin, Zhang, Agarwal, Slama, Ray, Schulman, Hilton, Kelton, Miller, Simens, Askell, Welinder, Christiano, Leike, and Lowe]{Ouyang2022TrainingLM}
Long Ouyang, Jeff Wu, Xu~Jiang, Diogo Almeida, Carroll~L. Wainwright, Pamela Mishkin, Chong Zhang, Sandhini Agarwal, Katarina Slama, Alex Ray, John Schulman, Jacob Hilton, Fraser Kelton, Luke Miller, Maddie Simens, Amanda Askell, Peter Welinder, Paul Christiano, Jan Leike, and Ryan Lowe.
\newblock Training language models to follow instructions with human feedback.
\newblock In \emph{Advances in Neural Information Processing Systems}, volume~35, 2022.
\newblock URL \url{https://proceedings.neurips.cc/paper/2022/hash/b1efde53be364a73914f58805a001731-Abstract-Conference.html}.

\bibitem[Paszke et~al.(2019)Paszke, Gross, Massa, Lerer, Bradbury, Chanan, Killeen, Lin, Gimelshein, Antiga, Desmaison, Köpf, Yang, DeVito, Raison, Tejani, Chilamkurthy, Steiner, Fang, Bai, and Chintala]{Paszke2019PyTorch}
Adam Paszke, Sam Gross, Francisco Massa, Adam Lerer, James Bradbury, Gregory Chanan, Trevor Killeen, Zeming Lin, Natalia Gimelshein, Luca Antiga, Alban Desmaison, Andreas Köpf, Edward Yang, Zach DeVito, Martin Raison, Alykhan Tejani, Sasank Chilamkurthy, Benoit Steiner, Lu~Fang, Junjie Bai, and Soumith Chintala.
\newblock {PyTorch}: An imperative style, high-performance deep learning library.
\newblock In \emph{Advances in Neural Information Processing Systems}, 2019.
\newblock URL \url{https://arxiv.org/abs/1912.01703}.

\bibitem[Qi et~al.(2024)Qi, Zeng, Xie, Chen, Jia, Mittal, and Henderson]{Qi2023FinetuningAL}
Xiangyu Qi, Yi~Zeng, Tinghao Xie, Pin-Yu Chen, Ruoxi Jia, Prateek Mittal, and Peter Henderson.
\newblock Fine-tuning aligned language models compromises safety, even when users do not intend to!
\newblock In \emph{International Conference on Learning Representations}, 2024.
\newblock URL \url{https://openreview.net/forum?id=hTEGyKf0dZ}.

\bibitem[Qian et~al.(2025)Qian, Zhou, Wang, Peng, Yi, Zhao, Huang, Xie, and Nie]{Qian2025Fino1OT}
Lingfei Qian, Weipeng Zhou, Yan Wang, Xueqing Peng, Han Yi, Yilun Zhao, Jimin Huang, Qianqian Xie, and Jian-Yun Nie.
\newblock {Fino1}: On the transferability of reasoning-enhanced {LLMs} and reinforcement learning to finance.
\newblock \emph{arXiv preprint arXiv:2502.08127}, 2025.
\newblock URL \url{https://arxiv.org/abs/2502.08127}.

\bibitem[Rafailov et~al.(2023)Rafailov, Sharma, Mitchell, Manning, Ermon, and Finn]{Rafailov2023DirectPO}
Rafael Rafailov, Archit Sharma, Eric Mitchell, Christopher~D. Manning, Stefano Ermon, and Chelsea Finn.
\newblock Direct preference optimization: Your language model is secretly a reward model.
\newblock In \emph{Advances in Neural Information Processing Systems}, volume~36, 2023.
\newblock URL \url{https://proceedings.neurips.cc/paper_files/paper/2023/hash/a85b405ed65c6477a4fe8302b5e06ce7-Abstract-Conference.html}.

\bibitem[Ren et~al.(2026)Ren, Wang, Cai, Shao, Guo, Xie, Li, Zhang, Hu, Shao, and Liu]{Ren2026MathCoT}
Qihan Ren, Peng Wang, Ruikun Cai, Shuai Shao, Dadi Guo, Yuejin Xie, Yafu Li, Quanshi Zhang, Xia Hu, Jing Shao, and Dongrui Liu.
\newblock Rethinking generalization in reasoning {SFT}: A conditional analysis on optimization, data, and model capability.
\newblock In \emph{Conference on Language Modeling}, 2026.
\newblock URL \url{https://arxiv.org/abs/2604.06628}.

\bibitem[Shao et~al.(2024)Shao, Wang, Zhu, Xu, Song, Bi, Zhang, Zhang, Li, Wu, and Guo]{Shao2024DeepSeekMathPT}
Zhihong Shao, Peiyi Wang, Qihao Zhu, Runxin Xu, Junxiao Song, Xiao Bi, Haowei Zhang, Mingchuan Zhang, Y.~K. Li, Y.~Wu, and Daya Guo.
\newblock {DeepSeekMath}: Pushing the limits of mathematical reasoning in open language models.
\newblock \emph{arXiv preprint arXiv:2402.03300}, 2024.
\newblock URL \url{https://arxiv.org/abs/2402.03300}.

\bibitem[Shenfeld et~al.(2025)Shenfeld, Pari, and Agrawal]{Shenfeld2025RLsRW}
Idan Shenfeld, Jyothish Pari, and Pulkit Agrawal.
\newblock {RL}'s razor: Why online reinforcement learning forgets less.
\newblock \emph{arXiv preprint arXiv:2509.04259}, 2025.
\newblock URL \url{https://arxiv.org/abs/2509.04259}.

\bibitem[Sheng et~al.(2024)Sheng, Zhang, Ye, Wu, Zhang, Zhang, Peng, Lin, and Wu]{Sheng2024HybridFlow}
Guangming Sheng, Chi Zhang, Zilingfeng Ye, Xibin Wu, Wang Zhang, Ru~Zhang, Yanghua Peng, Haibin Lin, and Chuan Wu.
\newblock {HybridFlow}: A flexible and efficient {RLHF} framework.
\newblock \emph{arXiv preprint arXiv:2409.19256}, 2024.
\newblock URL \url{https://arxiv.org/abs/2409.19256}.

\bibitem[Shuttleworth et~al.(2024)Shuttleworth, Andreas, Torralba, and Sharma]{Shuttleworth2024LoRAVF}
Reece Shuttleworth, Jacob Andreas, Antonio Torralba, and Pratyusha Sharma.
\newblock {LoRA} vs full fine-tuning: An illusion of equivalence.
\newblock \emph{arXiv preprint arXiv:2410.21228}, 2024.
\newblock URL \url{https://arxiv.org/abs/2410.21228}.

\bibitem[Sonthalia et~al.(2025)Sonthalia, Murray, and Mont{\'u}far]{Sonthalia2025LowRG}
Rishi Sonthalia, Michael Murray, and Guido Mont{\'u}far.
\newblock Low rank gradients and where to find them.
\newblock \emph{arXiv preprint arXiv:2510.01303}, 2025.
\newblock URL \url{https://arxiv.org/abs/2510.01303}.

\bibitem[Yan et~al.(2025)Yan, Zhang, and Tan]{Yan2025ERLoRAEG}
Weilong Yan, Xin Zhang, and Robby~T. Tan.
\newblock Er-{LoRA}: Effective-rank guided adaptation for weather-generalized depth estimation.
\newblock \emph{arXiv preprint arXiv:2509.00665}, 2025.
\newblock URL \url{https://arxiv.org/abs/2509.00665}.

\bibitem[Yang et~al.(2025)Yang, Li, Yang, Zhang, Hui, Zheng, Yu, Gao, Huang, Lv, Zheng, Liu, Zhou, Huang, Hu, Ge, Wei, Lin, Tang, Yang, Tu, Zhang, Yang, Yang, Zhou, Zhou, Lin, Dang, Bao, Yang, Yu, Deng, Li, Xue, Li, Zhang, Wang, Zhu, Men, Gao, Liu, Luo, Li, Tang, Yin, Ren, Wang, Zhang, Ren, Fan, Su, Zhang, Zhang, Wan, Liu, Wang, Cui, Zhang, Zhou, and Qiu]{Yang2025Qwen3TR}
An~Yang, Anfeng Li, Baosong Yang, Beichen Zhang, Binyuan Hui, Bo~Zheng, Bowen Yu, Chang Gao, Chengen Huang, Chenxu Lv, Chujie Zheng, Dayiheng Liu, Fan Zhou, Fei Huang, Feng Hu, Hao Ge, Haoran Wei, Huan Lin, Jialong Tang, Jian Yang, Jianhong Tu, Jianwei Zhang, Jianxin Yang, Jiaxi Yang, Jing Zhou, Jingren Zhou, Junyang Lin, Kai Dang, Keqin Bao, Kexin Yang, Le~Yu, Lianghao Deng, Mei Li, Mingfeng Xue, Mingze Li, Pei Zhang, Peng Wang, Qin Zhu, Rui Men, Ruize Gao, Shixuan Liu, Shuang Luo, Tianhao Li, Tianyi Tang, Wenbiao Yin, Xingzhang Ren, Xinyu Wang, Xinyu Zhang, Xuancheng Ren, Yang Fan, Yang Su, Yichang Zhang, Yinger Zhang, Yu~Wan, Yuqiong Liu, Zekun Wang, Zeyu Cui, Zhenru Zhang, Zhipeng Zhou, and Zihan Qiu.
\newblock {Qwen3} technical report.
\newblock \emph{arXiv preprint arXiv:2505.09388}, 2025.
\newblock URL \url{https://arxiv.org/abs/2505.09388}.

\bibitem[Zhang et~al.(2026)Zhang, Jin, Kang, and Wang]{Zhang2026PreferenceTA}
Peiyan Zhang, Haibo Jin, Liying Kang, and Haohan Wang.
\newblock Preference tuning as spectral update reorganization.
\newblock \emph{arXiv preprint arXiv:2607.20438}, 2026.
\newblock URL \url{https://arxiv.org/abs/2607.20438}.

\bibitem[Zhang et~al.(2025)Zhang, Li, Long, Zhang, Lin, Yang, Xie, Yang, Liu, Lin, Huang, and Zhou]{qwen3embedding}
Yanzhao Zhang, Mingxin Li, Dingkun Long, Xin Zhang, Huan Lin, Baosong Yang, Pengjun Xie, An~Yang, Dayiheng Liu, Junyang Lin, Fei Huang, and Jingren Zhou.
\newblock {Qwen3 Embedding}: Advancing text embedding and reranking through foundation models.
\newblock \emph{arXiv preprint arXiv:2506.05176}, 2025.
\newblock \doi{10.48550/arXiv.2506.05176}.
\newblock URL \url{https://arxiv.org/abs/2506.05176}.

\bibitem[Zhao et~al.(2023)Zhao, Gu, Varma, Luo, Huang, Xu, Wright, Shojanazeri, Ott, Shleifer, Desmaison, Balioglu, Damania, Nguyen, Chauhan, Hao, Mathews, and Li]{Zhao2023FSDP}
Yanli Zhao, Andrew Gu, Rohan Varma, Liang Luo, Chien-Chin Huang, Min Xu, Less Wright, Hamid Shojanazeri, Myle Ott, Sam Shleifer, Alban Desmaison, Can Balioglu, Pritam Damania, Bernard Nguyen, Geeta Chauhan, Yuchen Hao, Ajit Mathews, and Shen Li.
\newblock {PyTorch} {FSDP}: Experiences on scaling fully sharded data parallel.
\newblock \emph{arXiv preprint arXiv:2304.11277}, 2023.
\newblock URL \url{https://arxiv.org/abs/2304.11277}.

\bibitem[Zheng et~al.(2025)Zheng, Lu, Wang, Feng, Kuang, Xiong, and Zhang]{Zheng2025EasyR1}
Yaowei Zheng, Junting Lu, Shenzhi Wang, Zhangchi Feng, Dongdong Kuang, Yuwen Xiong, and Richong Zhang.
\newblock {EasyR1}: An efficient, scalable, multi-modality {RL} training framework.
\newblock GitHub repository, 2025.
\newblock URL \url{https://github.com/hiyouga/EasyR1}.

\bibitem[Zhu et~al.(2025)Zhu, Zhang, Huang, Su, Liu, Zhao, Fedorov, Pirsiavash, Sha, Lee, Pan, Wang, Tian, and Tai]{Zhu2025ThePN}
Hanqing Zhu, Zhenyu Zhang, Hanxian Huang, DiJia Su, Zechun Liu, Jiawei Zhao, Igor Fedorov, Hamed Pirsiavash, Zhizhou Sha, Jinwon Lee, David~Z. Pan, Zhangyang Wang, Yuandong Tian, and Kai~Sheng Tai.
\newblock The path not taken: {RLVR} provably learns off the principals.
\newblock \emph{arXiv preprint arXiv:2511.08567}, 2025.
\newblock URL \url{https://arxiv.org/abs/2511.08567}.

\bibitem[Zhu et~al.(2026)Zhu, Jin, Tu, Yuan, Bai, Liu, Li, and Zhao]{Zhu2026SFTCR}
Kejian Zhu, Zhuoran Jin, Shangqing Tu, Hongbang Yuan, Yushi Bai, Kang Liu, Juanzi Li, and Jun Zhao.
\newblock {SFT} conflicts, {RL} coexists: A theoretical and empirical analysis of multi-task learning for {LLMs}.
\newblock \emph{arXiv preprint arXiv:2608.03573}, 2026.
\newblock URL \url{https://arxiv.org/abs/2608.03573}.

\end{thebibliography}
\bibliographystyle{plainnat}
\endgroup

\clearpage
\appendix
\section{Paired Bootstrap for Retention and Domain Interaction}
\label{app:paired_bootstrap}

Subtracting the matched-domain gap from the far-domain gap gives
$\Delta_{\mathrm{DiD}}=22.73$ percentage points, with a 95\% interval of
$[12.45,33.51]$, as reported in Table~\ref{tab:retention_uncertainty}. This supports a larger
gap in the far condition for these checkpoints and evaluation questions.

\begin{table}[htbp]
\centering\small
\setlength{\tabcolsep}{4pt}
\caption{No-defense macro retention (\%) and GRPO--SFT gaps (percentage points), with 95\% paired-bootstrap percentile intervals in brackets.
DiD compares the far and matched gaps. Intervals are marginal and are not
adjusted for multiple comparisons.}
\label{tab:retention_uncertainty}
\begin{tabular}{@{}lccc@{}}
\toprule
Condition & SFT & GRPO & GRPO $-$ SFT \\
\midrule
Matched & 49.98 [39.38, 58.85] & 49.68 [40.87, 55.81] & -0.30 [-7.50, 6.16] \\
Near & 78.03 [68.21, 86.12] & 84.29 [74.25, 90.85] & 6.26 [-4.98, 15.41] \\
Far & 72.58 [62.15, 83.32] & 95.02 [87.21, 102.23] & 22.43 [13.12, 31.79] \\
\midrule
DiD (far $-$ matched) & \multicolumn{2}{c}{-} & 22.73 [12.45, 33.51] \\
\bottomrule
\end{tabular}
\end{table}

\paragraph{Evaluation records and pairing.}
The analysis uses eight evaluation conditions: the SFT and GRPO anchors,
and their no-defense endpoints after math-error (matched), code-error (near),
and XSum (far) interference. Each condition contains the same 500 MATH500
questions with four responses per question and the same 30 questions in each
AIME benchmark with sixteen responses per question. Exact question text is
used to align conditions, and reference answers agree across all eight.
Accuracy is the mean of the binary correctness scores across responses,
not pass@$k$. Point estimates use unrounded scores.

\paragraph{Resampling.}
We draw 10{,}000 bootstrap replicates with NumPy's PCG64 generator and seed 42. Within each benchmark, we sample its original number of questions with replacement. The same sampled indices are used for all eight conditions, while
different benchmarks are sampled independently. All generated responses to a sampled question are kept together. For every replicate, we recompute each benchmark's anchor and post-interference accuracy, divide by the matching
anchor accuracy, and average the three ratios with equal benchmark weight.
We then compute the GRPO-SFT gaps and the far-minus-matched DiD using those
same resampled quantities. All resampled anchor denominators were positive.
Intervals are the 2.5th and 97.5th percentiles, using linear interpolation.

\paragraph{Scope and interpretation.}
These intervals capture question-sampling uncertainty conditional on the
checkpoints and fixed generated responses. Anchor MacroRet is exactly 100\% by
definition in every replicate. The anchor's accuracy intervals are reported below.
Retention intervals can exceed 100\% and are not clipped. For completeness,
the near-minus-matched interaction is 6.56 percentage points with interval
$[-2.42,14.68]$. The primary DiD is the far-minus-matched contrast defined in
Equation~\ref{eq:did}. An interval containing zero is not an equivalence
result.

\begin{table}[htbp]
\centering\small
\setlength{\tabcolsep}{4pt}
\caption{Accuracy (\%) with 95\% paired question-bootstrap percentile
intervals. All interference rows use no defense.}
\label{tab:accuracy_uncertainty}
\begin{tabular}{@{}llccc@{}}
\toprule
Anchor & Condition & MATH500 & AIME24 & AIME25 \\
\midrule
SFT & Anchor & 81.55 [78.75, 84.20] & 23.96 [12.92, 36.67] & 25.42 [13.33, 39.17] \\
SFT & Matched & 61.15 [57.60, 64.60] & 8.33 [2.29, 15.83] & 10.21 [2.71, 20.00] \\
SFT & Near & 73.60 [70.45, 76.70] & 18.75 [8.33, 31.04] & 16.67 [7.29, 27.71] \\
SFT & Far & 79.10 [76.15, 82.10] & 14.79 [6.25, 24.79] & 15.00 [6.25, 25.21] \\
GRPO & Anchor & 85.40 [82.70, 88.00] & 24.17 [12.29, 37.29] & 21.67 [10.83, 34.17] \\
GRPO & Matched & 56.60 [53.55, 59.60] & 7.92 [2.08, 14.79] & 10.83 [3.54, 19.38] \\
GRPO & Near & 82.35 [79.60, 85.00] & 18.75 [8.33, 30.83] & 17.08 [6.46, 29.58] \\
GRPO & Far & 84.95 [82.35, 87.50] & 24.17 [12.08, 37.71] & 18.54 [8.33, 30.42] \\
\bottomrule
\end{tabular}
\end{table}

\section{Anchor Data and Training Details}
\label{app:anchor_details}

\subsection{Corpus selection and preprocessing}
Both anchors use the first 19{,}000 rows, in stored order, of the
20{,}480-row Math-CoT-20k corpus~\citep{Ren2026MathCoT}, available as
\path{jasonrqh/Math-CoT-20k} on Hugging Face. Its prompts are drawn from
OpenR1-Math-220k~\citep{OpenR1Math2025}, and its reference solutions are generated by
Qwen3-32B~\citep{Yang2025Qwen3TR}. The remaining 1{,}480 rows form a holdout pool, whose first
1{,}024 rows are used for math Fisher estimation. This is a deterministic
split, not random sampling: zero-based source rows 0-18{,}999 are training
rows and 19{,}000-20{,}023 are Fisher rows. Exact question-string comparison
finds no overlap between these two subsets, but semantic near-duplicates were not
checked. The SFT and GRPO conversion pipelines discard empty fields but retain
all 19{,}000 training rows. GRPO additionally filters prompts exceeding
2{,}048 tokens and shuffles with data seed 1.

\subsection{SFT optimization and completed epochs}
The SFT anchor uses the PyTorch~\citep{Paszke2019PyTorch} implementation of AdamW~\citep{Loshchilov2019AdamW} with learning rate
$5\times10^{-5}$, betas $(0.9,0.999)$, weight decay $0.01$, and gradient
clipping at norm 1. The global training batch size is 256 (32 per GPU across eight GPUs), with
a microbatch size of one per GPU. Sequences are right-truncated to a maximum length of
20{,}000 tokens. The training configuration specifies eight epochs and seed 1.
The reported anchor at step 592 completes eight epochs of 74 optimizer
updates each, processing 151{,}552 sample presentations. Each epoch uses
18{,}944 samples and drops the final incomplete batch of 56. The learning-rate schedule
is cosine, with a configured warmup ratio of $0.1$.

\subsection{GRPO optimization and rollout accounting}
The configured implementation uses token-averaged loss, asymmetric clipping
with $\epsilon_{\mathrm{low}}=0.2$ and $\epsilon_{\mathrm{high}}=0.3$
(a ratio interval of $[0.8,1.3]$), and a dual-clipping parameter of 3.
The KL penalty is retained as a loss term with fixed coefficient $\beta=0.01$
and a low-variance KL estimator. Online filtering is disabled.
For each prompt, $G=8$ responses are generated using vLLM~\citep{Kwon2023vLLM} with temperature
$1.0$, top-$p=1.0$, no top-$k$ restriction, and rollout seed 1. The prompt and
response length limits are 2{,}048 and 16{,}384 tokens, respectively.

The actor uses AdamW with learning rate $10^{-6}$, betas $(0.9,0.999)$,
weight decay $0.01$, a constant learning-rate schedule without warmup,
and maximum gradient norm 1. The rollout batch contains 128 prompts.
The configured actor global batch size is 64, with one PPO epoch per batch
and an update microbatch size of one per device. A GRPO step denotes one
rollout/update iteration, not one optimizer step. Step 410 corresponds
to 52{,}480 prompt presentations and 419{,}840 sampled responses, or
$410/148\approx2.7703$ dataloader epochs out of the configured maximum of three.

\subsection{Anchor update-token accounting}
\label{app:anchor_tokens}
We report cumulative response-update token counts at the selected anchor
checkpoints. Table~\ref{tab:anchor_tokens} distinguishes the recorded counts
from the GRPO stopping threshold.

\paragraph{SFT.}
The offline accounting procedure counts retained
response tokens after the sequence-length limit, excluding padding and
prompt tokens. For each sample presentation $i$ in optimizer step $t$, let
$L^{\mathrm{resp}}_{t,i}$ be the retained response length under the accounting
tokenization convention. Its cumulative counter is
\begin{equation}
T_{\oSFT}(s)=\sum_{t=1}^{s}\sum_{i\in\mathcal{B}_t}
\max\!\left(0,L^{\mathrm{resp}}_{t,i}-1\right).
\label{eq:sft_token_count}
\end{equation}
Repeated presentations across epochs are counted separately. At step 592,
the run has processed $592\times256=151{,}552$ presentations over eight
epochs, giving a cumulative offline count of
939{,}205{,}185.

\paragraph{GRPO.}
After each actor update, the training counter adds the number of valid
response-token positions in the entire update batch:
\begin{equation}
T_{\oGRPO}(s)=\sum_{t=1}^{s}\sum_{i,u}
m^{\mathrm{resp}}_{t,i,u},
\label{eq:tokens}
\end{equation}
where $t$ indexes rollout/update iterations, $i$ responses, and $u$ token
positions. The mask $m^{\mathrm{resp}}_{t,i,u}$ is one for a valid response-token
position and zero otherwise. Prompt and padding positions do not contribute. Responses with
zero advantage are included because the KL loss remains active
($\beta=0.01$). Online filtering is disabled. The counter is incremented once
per actor update and is not multiplied by the number of PPO epochs.
This run uses one PPO epoch.

The stopping threshold is 939{,}205{,}185 tokens. Since it is checked after
completing a batch update, the cumulative count recorded at step 410 is
942{,}891{,}140 tokens, exceeding the threshold by 3{,}685{,}955. Relative to
the SFT count, the difference is
\begin{equation}
100\frac{T_{\oGRPO}-T_{\oSFT}}{T_{\oSFT}}=0.3925\%.
\end{equation}

\begin{table}[htbp]
\centering\small
\caption{Anchor update-token accounting. The GRPO threshold is a stopping
criterion, not the final cumulative count.}
\label{tab:anchor_tokens}
\begin{tabular}{@{}llr@{}}
\toprule
Anchor & Quantity & Tokens \\
\midrule
SFT & Recorded offline count, step 592 & 939{,}205{,}185 \\
GRPO & Stopping threshold & 939{,}205{,}185 \\
GRPO & Recorded training count, step 410 & 942{,}891{,}140 \\
GRPO & Overshoot above threshold & 3{,}685{,}955 \\
\bottomrule
\end{tabular}
\end{table}

\paragraph{Counting convention.}
The SFT value above is the offline response-length-minus-one count used for
budget selection. Applying the current trainer's shifted loss mask to the
cached, untruncated sequences instead adds one target per presentation,
yielding $939{,}205{,}185+151{,}552=939{,}356{,}737$ tokens and a GRPO SFT
difference of 0.3763\%. This alternative is a reconstruction under the current
mask implementation, not an original training-log measurement. The reported
anchor comparison consistently uses the recorded counts in
Table~\ref{tab:anchor_tokens}.

\section{Implementation, Hyperparameters, and Runtime}
\label{app:implementation}

\begin{table}[htbp]
\centering\small
\setlength{\tabcolsep}{4pt}
\renewcommand{\arraystretch}{1.15}
\caption{Overview of the interference corpora used in the second training stage.
Distance refers to the relation between the interference domain and the anchored
math capability.}
\label{tab:interference}
\begin{tabularx}{\linewidth}{@{}>{\raggedright\arraybackslash}p{1.3cm}>{\raggedright\arraybackslash}p{2.7cm}>{\raggedright\arraybackslash}X>{\raggedleft\arraybackslash}p{1.3cm}>{\raggedright\arraybackslash}p{2.9cm}@{}}
\toprule
Tier & Corpus & Construction & Size & Anchor-domain relation \\
\midrule
Matched  & Sampled math-error trajectories  & Qwen3-0.6B rollouts on OpenR1-Math-220k, incorrect only & 7,603 & Same capability, wrong trajectories \\
Near     & Sampled code-error trajectories  & Qwen3-0.6B code trajectories on APPS                    & 4,580 & Adjacent structured-reasoning domain \\
Near     & APPS+ trajectories P / WF / WR   & Generated by deepseek-v4-flash. WF perturbs format and WR perturbs reasoning. & 2,890 & Adjacent domain with controlled corruption \\
Far      & XSum                             & Standard summarization corpus                          & 204,017 & Non-overlapping task and format \\
\bottomrule
\end{tabularx}
\end{table}

\begin{table}[htbp]
\centering\small
\caption{Hyperparameters and other implementation settings.
Interference settings are shared by both anchors unless stated otherwise.
Anchor-training and final-evaluation settings are reported separately.}
\label{tab:hyper}
\begin{tabular}{@{}llr@{}}
\toprule
Stage & Setting & Value \\
\midrule
Anchor (SFT)      & Checkpoint / steps            & 592 \\
Anchor (SFT)      & Optimizer / learning rate     & AdamW / $5\times10^{-5}$ \\
Anchor (SFT)      & LR schedule / warmup ratio    & Cosine / 0.1 \\
Anchor (SFT)      & Global / per-GPU microbatch   & 256 / 1 \\
Anchor (SFT)      & Max sequence length           & 20{,}000 \\
Anchor (SFT)      & Configured epochs / seed      & 8 / 1 \\
Anchor (\GRPO)    & Checkpoint / steps            & 410 \\
Anchor (\GRPO)    & Group size / PPO epochs       & 8 / 1 \\
Anchor (\GRPO)    & Clip low / high / dual        & 0.2 / 0.3 / 3 \\
Anchor (\GRPO)    & KL loss coefficient           & 0.01 (enabled) \\
Anchor (\GRPO)    & Optimizer / learning rate     & AdamW / $10^{-6}$ \\
Anchor (\GRPO)    & Rollout / actor global batch  & 128 / 64 \\
Anchor (\GRPO)    & Rollout temperature / top-$p$ & 1.0 / 1.0 \\
Anchor (\GRPO)    & Prompt / response limit       & 2{,}048 / 16{,}384 \\
Anchor (\GRPO)    & Data / rollout seed           & 1 / 1 \\
Anchor (\GRPO)    & Response-token stopping threshold & 939{,}205{,}185 \\
Anchor (both)     & Training samples              & 19{,}000 (Math-CoT-20k subset) \\
Anchor (both)     & Recorded token counts (see text) & SFT 939.21M, \GRPO\ 942.89M \\
\addlinespace
Interference (matched, near) & Learning rate & $1\times10^{-5}$ \\
Interference (matched, near) & Batch size    & 128 \\
Interference (matched, near) & Max sequence length & 16{,}384 \\
Interference (matched, near) & Steps         & 10 \\
Interference (far)           & Batch size    & 2{,}000 \\
\addlinespace
EWC penalty      & Diagonal Fisher source & 1{,}024 held-out Math-CoT-20k samples \\
EWC penalty      & Strength $\lambda$     & $\{10^{6},10^{7}\}$ \\
Frozen variant   & Frozen layers (zero-based) & 32-34 \\
Freezing comparison & Training seed (both variants) & 42 \\
\addlinespace
Evaluation       & MATH500                & 500 problems, 4 response each \\
Evaluation       & AIME24 / AIME25        & 30 problems each, 16 responses each \\
Evaluation       & Temperature / top-$p$  & $0.6$ / $0.95$ \\
Evaluation       & Top-$k$                & $-1$ \\
Evaluation       & Max response length    & 32{,}768 tokens \\
Evaluation       & Seed                   & 42 \\
\bottomrule
\end{tabular}
\end{table}

\paragraph{Implementation and evaluation.}
\label{sec:impl}
Both anchors use full-parameter training and are trained to the checkpoints
given in Table~\ref{tab:hyper}.
Their recorded update token counts differ by 0.39\%. All
experiments are conducted using the \texttt{verl} framework~\citep{Sheng2024HybridFlow}. The \GRPO{}
anchor uses EasyR1~\citep{Zheng2025EasyR1} on top of \texttt{verl}. Its configuration uses one node
with eight NVIDIA A100 GPUs (80 GB each), FSDP~\citep{Zhao2023FSDP}, bfloat16 parameter precision,
gradient checkpointing,
and actor parameter and optimiser offloading. The SFT anchor is trained on
one node with eight NVIDIA A100 GPUs (80 GB each) using FSDP2~\citep{PyTorchFSDP2}, bfloat16 model
precision, gradient
checkpointing, padding removal, and no CPU or parameter offloading. LoRA is
disabled for both anchors. All interference runs use the same \texttt{verl}
framework and AdamW optimiser as SFT anchoring, with the interference learning
rate specified in Table~\ref{tab:hyper}. Each configuration is trained for
10 steps on a single node, using either eight NVIDIA A100 GPUs (80 GB each)
or, for some interference runs, four NVIDIA A800 GPUs (80 GB each).
Training the SFT and \GRPO{} anchors to steps 592 and 410 takes 25 and
47 hours, respectively. On eight GPUs, the ten-step math and code interference
runs finish within 20 minutes, and XSum interference runs finish within two
hours. These runtimes include evaluation, generation, and checkpoint saving.
Math evaluation uses four responses per MATH500 problem and sixteen per AIME
problem, with temperature $0.6$, top-$p=0.95$, a 32{,}768-token response
limit, and seed 42. Math answers are checked with
Math-Verify~\citep{Kydlicek2025MathVerify}, and code benchmarks use
EvalPlus~\citep{Liu2023EvalPlus}.
For each anchor, the math Fisher diagonal is estimated from its own greedy
responses to the first 1{,}024 held-out prompts, selected in stored order
without random sampling, and reused across its EWC runs. Final evaluation uses seed 42, while SFT
anchor training, \GRPO{} data shuffling, and \GRPO{} rollouts use seed 1.
The frozen and unfrozen runs in the layer-freezing comparison both use
training seed 42 and identical data order.

\section{Benchmark Trajectories under No-Defense Interference}
\label{app:interference_trajectories}

\begin{figure}[htbp]
\centering
\includegraphics[width=\linewidth]{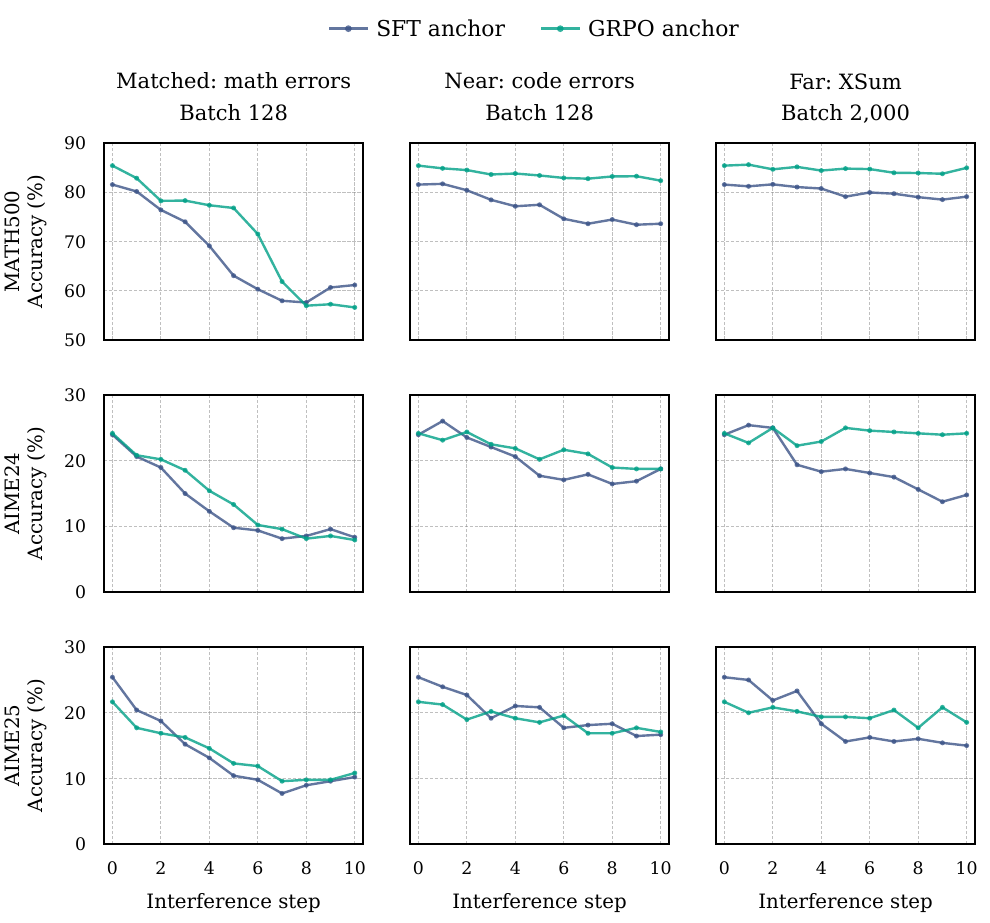}
\caption{Benchmark accuracy (\%) over ten steps of no-defense interference
for the SFT and GRPO anchors on Qwen3-4B. Columns show interference tiers
and their batch sizes. Rows show MATH500, AIME24, and AIME25, with a common
vertical scale across tiers for each benchmark. Markers denote recorded
evaluations at every step from 0 to 10, including the pre-interference anchor
at step 0. Lines connect consecutive recorded evaluations.}
\label{fig:interference_benchmarks}
\end{figure}

Figure~\ref{fig:interference_benchmarks} reports the stepwise math benchmark
accuracies for the six Qwen3-4B interference runs: each of the SFT and GRPO
anchors continued on matched-domain math errors, near-domain APPS code
errors, or far-domain XSum summaries, without a defense. The batch sizes in
Table~\ref{tab:tiers} are chosen so that the ten-step probe can produce
observable interference effects. The recorded trajectories show how these
effects develop on each benchmark. Step 0 is the corresponding anchor
evaluation, while steps 1-10 are interference updates. Accuracy averages four
responses per MATH500 problem and sixteen per AIME problem, using the
evaluation settings in Appendix~\ref{app:implementation}. All six runs have
recorded evaluations at every step from 0 to 10.

\section{On-Policy SFT: Capability Retention and Generation Behavior}
\label{app:on_policy_sft}

\paragraph{Construction of the control.}
To examine whether GRPO-generated training data can reproduce the retention
pattern under a supervised objective, we construct an additional Qwen3-4B
anchor. We sample five responses from the GRPO model for each prompt in the
same 19{,}000-example Math-CoT training subset used by the original anchors,
retain at most one correct trajectory per prompt, and discard prompts with
no correct response, yielding 15{,}872 trajectories for training.
Starting again from the common base model, we train
on these trajectories with SFT for eight epochs, obtaining the step-496
anchor. We call this control \emph{on-policy SFT} to denote the source of
its supervision: the trajectories come from the GRPO policy, and are not
refreshed online from the evolving SFT student. We then evaluate the anchor
under matched-domain math-error, near-domain code-error, and far-domain
XSum interference. This is a partial control for the source of the training
targets. The token matching established for the original SFT/GRPO pair
does not establish matched exposure for this additional anchor.

\begin{table}[htbp]
\centering\small
\setlength{\tabcolsep}{5pt}
\caption{On-policy SFT under three interference domains on Qwen3-4B.
Accuracy and MacroRet are percentages. MacroRet follows Equation~\ref{eq:ret},
with each benchmark accuracy normalized by the corresponding accuracy of the
model's own anchor. Original SFT and GRPO
no-defense results are repeated for comparison.}
\label{tab:on_policy_sft_retention}
\begin{tabular*}{\linewidth}{@{\extracolsep{\fill}}llrrrr@{}}
\toprule
Condition & Anchor & MATH500 & AIME24 & AIME25 & MacroRet \\
\midrule
No interference & SFT & 81.55 & 23.96 & 25.42 & 100.00 \\
 & GRPO & 85.40 & 24.17 & 21.67 & 100.00 \\
 & On-policy SFT & 81.00 & 25.21 & 22.71 & 100.00 \\
\midrule
Matched: math errors & SFT & 61.15 & 8.33 & 10.21 & 49.98 \\
 & GRPO & 56.60 & 7.92 & 10.83 & 49.68 \\
 & On-policy SFT & 55.40 & 6.67 & 8.33 & 43.84 \\
\midrule
Near: code errors & SFT & 73.60 & 18.75 & 16.67 & 78.03 \\
 & GRPO & 82.35 & 18.75 & 17.08 & 84.29 \\
 & On-policy SFT & 75.25 & 20.63 & 19.17 & 86.38 \\
\midrule
Far: XSum & SFT & 79.10 & 14.79 & 15.00 & 72.58 \\
 & GRPO & 84.95 & 24.17 & 18.54 & 95.02 \\
 & On-policy SFT & 79.90 & 22.08 & 20.42 & 92.05 \\
\bottomrule
\end{tabular*}
\end{table}

\paragraph{Cross-domain retention approaches GRPO.}
Table~\ref{tab:on_policy_sft_retention} shows that on-policy SFT reaches
86.38\% MacroRet under near-domain interference and 92.05\% under
far-domain interference, compared with 84.29\% and 95.02\% for GRPO.
Its margins over ordinary SFT are 8.35 and 19.47 percentage points,
respectively. Thus, a supervised objective trained on correct
GRPO-generated trajectories can recover much of the cross-domain retention
observed with GRPO. This is consistent with a contribution from the training
targets and their selection, rather than attributing the entire difference
to the GRPO objective. Similar retention ratios do not imply identical
absolute accuracy: under near-domain interference, for example, MATH500
accuracy remains lower for on-policy SFT than for GRPO (75.25\% versus
82.35\%). Under matched-domain interference, on-policy SFT instead falls to
43.84\% MacroRet, below the point estimates for both original anchors.

\paragraph{Response length and truncation.}
We analyze MATH500 response lengths and truncation rates for twelve checkpoints in
Table~\ref{tab:on_policy_sft_lengths}. Evaluation uses temperature 0.6,
top-$p=0.95$, no top-$k$ restriction, seed 42, and a response limit of
32{,}768 tokens. The length summaries contain 2{,}000 responses per
checkpoint.

\begin{table}[htbp]
\centering\small
\setlength{\tabcolsep}{4pt}
\caption{MATH500 accuracy, response length, and truncation under three
interference domains. Mean and median lengths are in tokens, while
accuracy and truncation rates are percentages.}
\label{tab:on_policy_sft_lengths}
\begin{tabular*}{\linewidth}{@{\extracolsep{\fill}}lrrrr@{}}
\toprule
Anchor & Acc. (\%) & Mean length & Median & Truncated (\%) \\
\midrule
\multicolumn{5}{@{}c@{}}{\textit{Before interference}} \\
SFT & 81.55 & 8{,}426.9 & 3{,}424 & 14.45 \\
GRPO & 85.40 & 3{,}108.7 & 1{,}152 & 2.95 \\
On-policy SFT & 81.00 & 11{,}958.8 & 1{,}828 & 31.50 \\
\midrule
\multicolumn{5}{@{}c@{}}{\textit{After matched interference}} \\
SFT & 61.15 & 13{,}070.0 & 2{,}424 & 36.20 \\
GRPO & 56.60 & 21{,}118.9 & 32{,}634 & 59.60 \\
On-policy SFT & 55.40 & 25{,}149.5 & 32{,}675 & 74.60 \\
\midrule
\multicolumn{5}{@{}c@{}}{\textit{After near-domain interference: code}} \\
SFT & 73.60 & 10{,}794.5 & 3{,}338 & 25.60 \\
GRPO & 82.35 & 3{,}570.0 & 911 & 6.05 \\
On-policy SFT & 75.25 & 3{,}423.7 & 1{,}154 & 4.50 \\
\midrule
\multicolumn{5}{@{}c@{}}{\textit{After far-domain interference: XSum}} \\
SFT & 79.10 & 3{,}070.1 & 553 & 6.95 \\
GRPO & 84.95 & 2{,}498.1 & 818 & 3.00 \\
On-policy SFT & 79.90 & 4{,}528.5 & 1{,}278 & 7.95 \\
\bottomrule
\end{tabular*}
\end{table}

\paragraph{Matched interference amplifies a pre-existing termination instability.}
The on-policy SFT anchor is already unusually prone to very long generations before interference. Its median response length (1,828 tokens) is below that of ordinary SFT (3,424), yet its mean is substantially larger (11,958.8 versus 8,426.9), and 31.50\% of responses reach the 32,768-token limit. This combination indicates a heavy upper tail rather than a uniform shift toward longer responses. Matched-domain interference strongly amplifies this pre-existing instability: the truncation rate rises to 74.60\%, the mean length more than doubles, and the median itself moves to 32,675 tokens. Thus, after matched interference, excessive continuation becomes the
dominant rather than tail behavior. Qualitative inspection shows that
some of these long responses continue generating repetitive reasoning or
additional boxed answers even after an extractable answer has already been
produced.

\paragraph{Cross-domain interference has a different generation profile.}
Under code and XSum interference, on-policy SFT mean response
lengths are 3{,}423.7 and 4{,}528.5 tokens, with truncation rates of 4.50\%
and 7.95\%, respectively. Both are substantially below the anchor's
11{,}958.8 tokens and 31.50\% truncation, in contrast to the deterioration
after matched interference. For the original anchors, code interference
also separates the generation profiles: ordinary SFT averages 10{,}794.5
tokens with 25.60\% truncation, compared with 3{,}570.0 tokens and 6.05\%
for GRPO. Under XSum, both original anchors generate shorter responses,
with truncation rates of 6.95\% for SFT and 3.00\% for GRPO. Thus, the
pronounced generation failure under matched interference does not extend
uniformly to the cross-domain evaluations.

\paragraph{Accuracy and termination reveal distinct dimensions of degradation.}
The matched-domain endpoint degrades along two separable behavioral
dimensions. First, MATH500 accuracy drops from 81.00\% before interference
to 55.40\% after matched-domain interference. Second, generation completion
deteriorates sharply: the truncation rate rises from 31.50\% to 74.60\%,
with most responses reaching the 32{,}768-token decoding limit. These two
quantities are related but are not interchangeable.

Among the 1{,}492 truncated responses after matched interference, 43.16\%
are still scored correct under first-box extraction, whereas 91.34\% of the
508 non-truncated responses are correct. These conditional accuracies recover
the reported overall MATH500 accuracy:
\[
0.746\times43.16\% + 0.254\times91.34\% \approx 55.40\%.
\]
Thus, truncation is not merely a cosmetic decoding artifact: it co-occurs
with substantial correctness degradation. At the same time, the non-trivial
accuracy among truncated generations shows that answer extraction can
partially mask severe generation-control failures. In particular, a response
may produce a correct early \texttt{\textbackslash boxed\{\}} answer and
still continue for many thousands of tokens with repetitive reasoning or
additional boxed answers.

We therefore treat answer correctness and successful termination as distinct
behavioral measurements. The former captures whether the model produces a
correct extractable answer, while the latter captures whether generation
finishes within the decoding budget. The present statistics do not establish
a causal direction between the two: difficult problems may induce both longer
responses and lower accuracy, and truncation itself may also prevent otherwise
valid generations from terminating cleanly.

\paragraph{Interpretation and scope.}
These results refine the interpretation of the original SFT--GRPO comparison. Replacing the original supervised targets with correct trajectories sampled from the GRPO policy substantially closes the cross-domain retention gap, suggesting that the source and selection of training trajectories contribute materially to the retention advantage and that the original difference cannot be attributed to the GRPO objective alone. At the same time, similar cross-domain retention does not imply similar generation behavior: the on-policy SFT anchor already exhibits a heavier long-response tail before interference, and matched-domain training amplifies this instability until most MATH500 responses reach the decoding limit. A plausible explanation is that same-domain SFT updates destabilize the decision to terminate after a completed solution, but the present experiments do not directly measure EOS probabilities or establish a causal mechanism. We therefore interpret the length and truncation results as evidence of generation-control degradation rather than direct evidence of EOS failure. Overall, GRPO-generated supervision is sufficient to reproduce much of GRPO's cross-domain retention under SFT, but it is not sufficient to reproduce its full generation profile under matched-domain interference.

\section{Numerical Details of Parameter-Space Diagnostics}
\label{app:geometry_details}

\begin{figure}[htbp]
\centering
\includegraphics[width=\columnwidth,trim=12bp 0bp 12bp 0bp,clip]{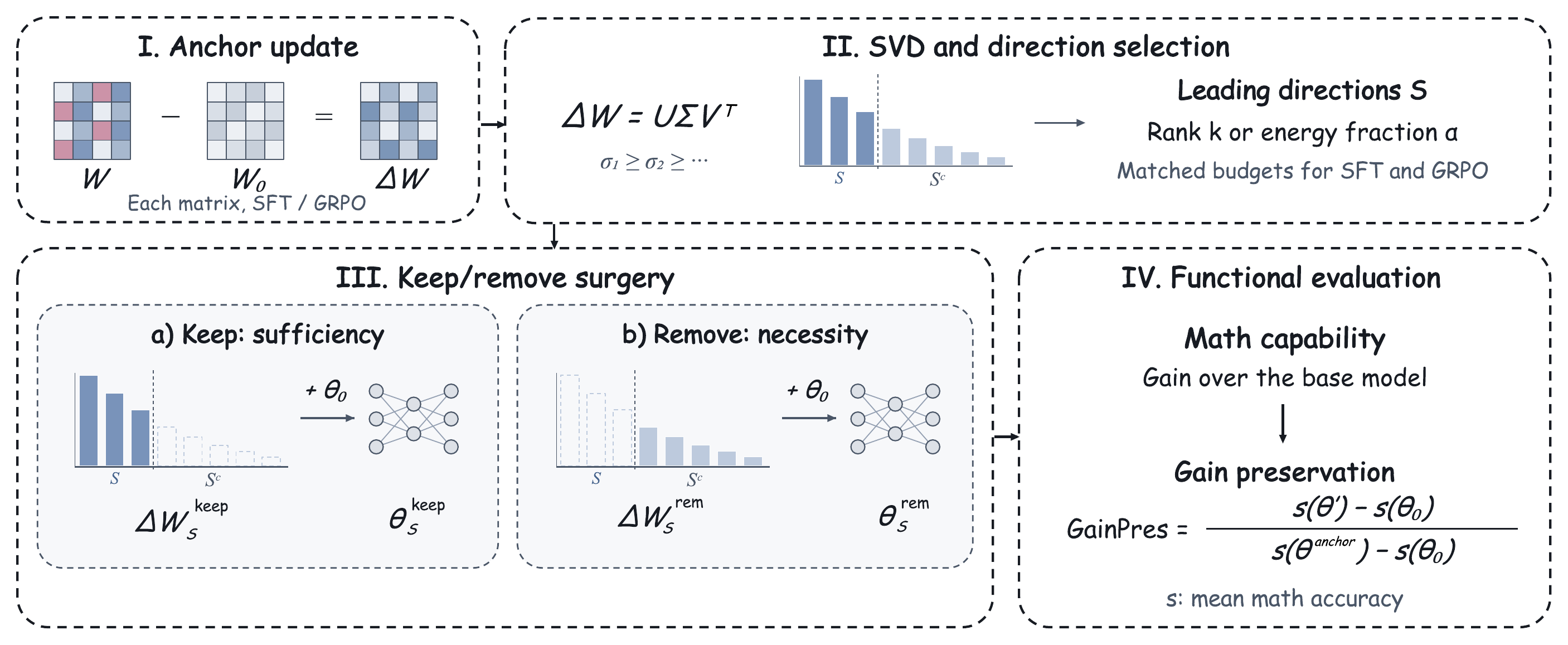}
\caption{Spectral surgery. The same budget is used in both variants so that
sufficiency (keep) and necessity (remove) are measured at matched cost.}
\label{fig:surgery}
\end{figure}

\begin{table}[htbp]
\centering\small
\caption{Static parameter-space statistics of the two anchor updates,
$\Delta W_{\mathrm{SFT}}$ and $\Delta W_{\GRPO}$, relative to the same base
model. These are descriptive statistics rather than competition metrics.
Bold marks lower effective rank and rank budget, or larger spectral energy,
Fisher-coordinate energy share, and profile distance, as applicable. These
quantities describe different aspects of the updates.}
\label{tab:geometry}
\setlength{\tabcolsep}{4pt}
\begin{tabular}{@{}>{\raggedright\arraybackslash}p{3.2cm}>{\raggedright\arraybackslash}p{6.2cm}rr@{}}
\toprule
Quantity & Interpretation & SFT & \GRPO \\
\midrule
Global cosine between the two updates & Direction agreement over 252 projection matrices (one number for the pair) & \multicolumn{2}{c}{0.003064} \\
Normalized effective rank & How many directions the update spreads over (smaller values mean greater concentration) & 0.71976 & \textbf{0.24950} \\
Top-1 singular energy (share) & Fraction of update energy carried by the leading direction (larger values mean greater concentration) & 0.0106386 & \textbf{0.0672467} \\
Directions to reach 90\% of update energy & Compressibility (fewer directions indicate greater compressibility) & 1656.46 & \textbf{981.57} \\
Top 1\% math-Fisher update energy (\%) & Share of squared update norm on globally ranked high-importance coordinates & 0.6801 & \textbf{1.2338} \\
Fr\'echet distance, math vs.\ code Fisher & Distance between normalized diagonal importance profiles & 0.5952 & \textbf{0.7336} \\
\bottomrule
\end{tabular}
\end{table}

\subsection{Per-matrix spectral selection and aggregation}
\label{app:spectral_aggregation}
The spectral diagnostics and interventions use $M=252$ projection matrices:
\texttt{q/k/v/o\_proj} and \texttt{gate/up/down\_proj} in each of 36 transformer
layers. For each anchor, we independently decompose each target update as
$\Delta W_m=W_m-W_{0,m}=U_m\Sigma_mV_m^\top$, with singular values in descending
order, rank $r_m$, and energy $e_m=\|\Delta W_m\|_F^2=\sum_i\sigma_{m,i}^2$.
All selections below are made within each matrix. Singular components are
not pooled across matrices before truncation.

\paragraph{Selection and component counts.}
For a Top-$k$ budget, each matrix selects its own leading components
$S_m=\{1,\ldots,k_m\}$, where $k_m=\min(k,r_m)$. For an energy budget
$\alpha$, the selected count is instead
\begin{equation}
k_{m,\alpha}=\min\left\{j:\frac{\sum_{i=1}^{j}\sigma_{m,i}^2}{e_m}
\geq\alpha\right\},\qquad S_m=\{1,\ldots,k_{m,\alpha}\}.
\label{eq:matrix_energy_budget}
\end{equation}
Both anchors share the same $k$ or $\alpha$, but each anchor and matrix has
its own selected set $S_m$. Equation~\ref{eq:surgery} is applied separately
to every target matrix using this set. Non-target tensors retain their parent
weights, subject to export precision.

The total number of selected rank-one components is
\begin{equation}
N(S)=\sum_{m=1}^{M}|S_m|.
\label{eq:spectral_component_count}
\end{equation}
Thus Top-1, Top-8, and Top-32 select exactly 252, 2,016, and 8,064 components,
respectively, for either anchor. In particular, Top-1 means one component
\emph{per target matrix}, not one component for the entire model. The target
matrices contain 534,528 available singular components in total. The keep intervention retains
$N(S)$ selected components, and the remove intervention retains the complementary
$534{,}528-N(S)$ components. For energy budgets, $N(S)=\sum_m k_{m,\alpha}$
and need not agree between anchors. Matching an energy threshold therefore
does not match the total component count. The exported weights remain dense
and retain their original shapes.

\paragraph{Energy aggregation.}
The ``Energy carried'' columns of Table~\ref{tab:spectral} report $100E(S)$,
where
\begin{equation}
E(S)=\frac{\sum_{m=1}^{M}\sum_{i\in S_m}\sigma_{m,i}^2}
{\sum_{m=1}^{M}e_m}
=\sum_{m=1}^{M}a_m E_m(S_m).
\label{eq:spectral_energy_aggregation}
\end{equation}
Here $a_m=e_m/\sum_n e_n$ and
$E_m(S_m)=\sum_{i\in S_m}\sigma_{m,i}^2/e_m$.
This is a ratio of summed energies, equivalently an update-energy-weighted
mean of per-matrix energy shares, rather than an unweighted matrix mean.
It measures the fraction kept in the keep intervention or deleted in the
remove intervention. The denominator includes only the 252 target matrices,
excluding embeddings, the output head, norms, biases, and other non-target
tensors. Their total update energies are 729.8323 for SFT and 1.97670 for
GRPO. The absolute selected energy is $E(S)\sum_m e_m$. A per-matrix energy
threshold can be exceeded because singular components are indivisible, so
the measured global share may be slightly larger than the nominal budget.

\paragraph{Aggregation of static spectral statistics.}
The spectral rows of Table~\ref{tab:geometry} use the same energy weights,
computed separately for each anchor. With $p_{m,i}=\sigma_{m,i}^2/e_m$ and
$H(p_m)=-\sum_i p_{m,i}\log p_{m,i}$, the reported quantities are
\begin{equation}
\overline R=\sum_m a_m\frac{\exp H(p_m)}{r_m},\qquad
E_k=\frac{\sum_m\sum_{i=1}^{\min(k,r_m)}\sigma_{m,i}^2}{\sum_m e_m},\qquad
\overline k_\alpha=\sum_m a_m k_{m,\alpha}.
\label{eq:spectral_statistic_aggregation}
\end{equation}
Thus the normalized effective rank is a weighted mean of normalized
per-matrix ranks, not the effective rank of a pooled spectrum. The reported
90\%-energy counts, 1656.46 for SFT and 981.57 for GRPO, are weighted means
of per-matrix counts and can be noninteger. Their ratio is approximately
59\%. This is not a ratio of total component counts or storage costs. The
Top-1 energy share, in contrast, directly measures the global target-update
energy captured by the 252 leading per-matrix components.

\subsection{Update alignment and analytic randomization baselines}
The FP32-anchor result is $0.0030641408$, while the separate matrix mean is
$0.0038311391$. The common base checkpoint is BF16, while both anchors are
FP32. Differences, products, and matrix reductions in the main calculation
use FP32, followed by double-precision scalar accumulation.

For a coordinate-randomized baseline, we fix the SFT update and independently
permute the GRPO coordinates within each matrix. This preserves each matrix's
shape, values, and Frobenius norm. Analytic moments give mean
$-1.1763\times10^{-8}$ and standard deviation $1.66734\times10^{-5}$.
The normal-approximation central 95\% null interval is
$[-3.26911,3.26676]\times10^{-5}$. The observed cosine has
$z=183.7744$ and approximate two-sided $\log_{10}p=-7336.0862$.
These are analytic null calculations, not finite random-draw estimates or a
confidence interval for the observed cosine. An independent sign-flip null
gives mean zero and standard deviation $1.66876\times10^{-5}$.
Thus the update alignment is geometrically small but distinguishable from
coordinate-randomized directions.

\subsection{Fisher profile preprocessing and parameter scope}
Both the SFT step-592 and GRPO step-410 anchors use FP32 parameter
precision for all math- and code-domain Fisher analyses, including Fisher
profile distances and update-energy projections onto high-Fisher coordinates.

The distance uses raw Fisher estimates before the maximum-based
normalization applied to EWC weights. Before global sum normalization, NaN and infinite
entries are set to zero and negative entries are clamped to zero. Estimation
aborts on a nonfinite loss, and distance computation rejects a nonpositive
or nonfinite normalizing sum. All 398 parameter tensors, comprising
4{,}022{,}468{,}096 coordinates, are included with strict name/shape matching.
Chunk calculations use FP32 with FP64 accumulation. This parameter scope is
larger than that used for update cosine.

\subsection{Update energy in high-Fisher coordinates}
\label{app:fisher_projection}
We compute $\Delta\theta_o=\theta_o-\theta_0$ for the SFT step-592 and
GRPO step-410 anchors relative to the same Qwen3-4B model. Each anchor's
math and code rankings use its own raw diagonal Fisher, estimated from
1{,}024 self-generated responses per domain at the corresponding anchor.

\paragraph{Parameter scope and selection.}
All four comparisons use the same 398 independent parameter tensors,
containing $P=4{,}022{,}468{,}096$ coordinates. Fisher values are ranked
globally rather than separately within tensors. The 0.1\%, 1\%, 5\%, and
10\% budgets select 4{,}022{,}469, 40{,}224{,}681, 201{,}123{,}405, and
402{,}246{,}810 coordinates, respectively. Ties are resolved by stored
parameter order so that exactly $\lceil qP\rceil$ coordinates are selected.
The math and code rankings are independent, but share the same update-energy
denominator within each anchor: $\|\Delta\theta_{\oSFT}\|_2^2=773.7584122945$
and $\|\Delta\theta_{\oGRPO}\|_2^2=2.0867763306$. Squared energies are
accumulated in double precision. The statistic is a fraction of each anchor's
own update energy. It does not compare absolute displacement magnitudes.

\paragraph{Random-coordinate reference.}
For a uniformly selected subset $S$ of $k$ coordinates, each coordinate has
inclusion probability $k/P$. For any fixed nonzero update,
\begin{equation}
\mathbb{E}_{S}\!\left[
\frac{\sum_{j\in S}(\Delta\theta_j)^2}{\|\Delta\theta\|_2^2}
\right]=\frac{k}{P}.
\end{equation}
The reference in Table~\ref{tab:fisher_projection} is this analytic
expectation, not an empirical null interval. SFT is below and GRPO above
the reference at all four tested budgets in both domains. No uncertainty
interval or hypothesis test is inferred from this ordering.

\section{Recorded Layer-Freezing Trajectories}
\label{app:freeze_results}
Table~\ref{tab:freeze_steps} gives the values plotted in
Figure~\ref{fig:freeze}. Accuracy averages all four responses per problem
on 500 MATH500 problems (2{,}000 responses per checkpoint). This metric is not pass@4.
Step 0 is the existing anchor evaluation, shared by the corresponding frozen
and unfrozen curves. Steps 1-10 are the recorded run summaries.
For each anchor, both variants use training seed 42, identical data order,
and identical training configurations apart from freezing all parameters in
zero-based transformer layers 32-34 (the 33rd-35th blocks).

\begin{table}[htbp]
\centering\small
\caption{MATH500 average accuracy (\%) during APPS code-error interference.
Frozen runs freeze all parameters in zero-based transformer layers 32-34.}
\label{tab:freeze_steps}
\begin{tabular}{@{}rcccc@{}}
\toprule
 & \multicolumn{2}{c}{SFT anchor} & \multicolumn{2}{c}{GRPO anchor} \\
\cmidrule(lr){2-3}\cmidrule(l){4-5}
Step & Unfrozen & Frozen & Unfrozen & Frozen \\
\midrule
0 & 81.55 & 81.55 & 85.40 & 85.40 \\
1 & 81.70 & 81.60 & 84.85 & 86.05 \\
2 & 80.40 & 81.90 & 84.50 & 84.45 \\
3 & 78.45 & 79.85 & 83.60 & 83.70 \\
4 & 77.15 & 78.10 & 83.80 & 83.60 \\
5 & 77.45 & 77.50 & 83.40 & 83.65 \\
6 & 74.60 & 75.00 & 82.90 & 83.80 \\
7 & 73.60 & 74.95 & 82.75 & 83.45 \\
8 & 74.45 & 74.20 & 83.20 & 83.80 \\
9 & 73.40 & 73.45 & 83.25 & 82.40 \\
10 & 73.60 & 73.05 & 82.35 & 81.90 \\
\bottomrule
\end{tabular}
\end{table}

\section{Prompt-Domain Distance Analysis}
\label{app:domain_distance}
\paragraph{Data and preprocessing.}
The common reference is the first 1{,}024 MathCoT holdout questions
(source rows 19{,}000-20{,}023), identical to the prompts used for math Fisher
estimation. For each other group, we sample 1{,}024 unique cleaned prompts
without replacement with NumPy seed 20260919, excluding empty prompts and
exact overlap with the reference. The MathCoT internal baseline is sampled
from rows 0-18{,}999. It compares the stored training and holdout portions,
not two random halves of the holdout, and may reflect source ordering.
The interference groups use the math-error, code-error, controlled APPS,
and XSum corpora in Table~\ref{tab:interference}. Sampling is at the corpus
level, independently of historical continuation batches.

For chat-formatted inputs we extract the user message and remove the
assistant prefix. We remove the known math answer-format suffix, both
Standard Input and Call-Based code instructions, the controlled APPS task
prefix, and the XSum summarization instruction. Problem statements,
input/output examples, function signatures, docstrings, and article contents
are retained. Response targets are not embedded. The entire controlled APPS
P/WF/WR prompt lists are identical in order, so all three conditions use the
same sampled embeddings. Sample identities are tracked using source row
indices and exact prompt text.

\begin{figure}[htbp]
\centering
\includegraphics[width=\linewidth]{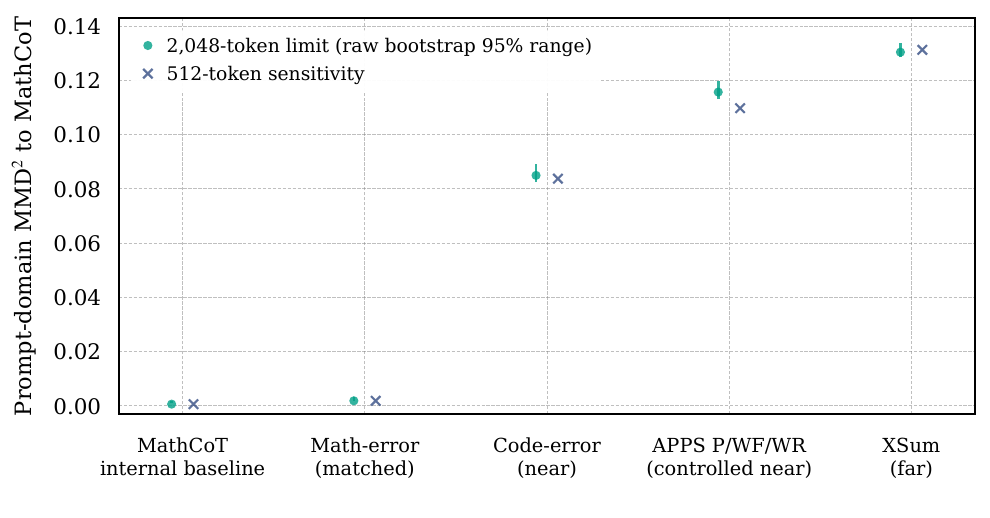}
\caption{Prompt-domain MMD$^2$ relative to MathCoT. Teal circles use a
2{,}048-token limit. Vertical segments show raw 2.5th--97.5th bootstrap
percentile ranges, subject to the near-null bias discussed above, rather
than calibrated confidence intervals. Navy crosses show the 512-token
sensitivity analysis. The vertical axis is linear. A common bandwidth
is used across conditions within each input-length setting.}
\label{fig:domain_distance}
\end{figure}

\paragraph{Encoder and statistic.}
We use the frozen Qwen3-Embedding-0.6B checkpoint~\citep{qwen3embedding} at revision
\path{97b0c614be4d77ee51c0cef4e5f07c00f9eb65b3}, with no retrieval instruction.
Inference uses BF16, left padding, right truncation, the last non-padding
hidden state, and FP32 L2 normalization. The resulting vectors have 1{,}024
dimensions. Kernel computations and reductions use FP64.
For two sets of unit vectors $X=\{x_i\}_{i=1}^n$ and
$Y=\{y_j\}_{j=1}^n$, where $n=1{,}024$, we compute the biased squared
maximum mean discrepancy (MMD) estimator~\citep{Gretton2012MMD}:
\begin{equation}
\begin{aligned}
\widehat{\mathrm{MMD}}^2_V(X,Y)
&=\frac{1}{n^2}\sum_{i,i'} k(x_i,x_{i'})
+\frac{1}{n^2}\sum_{j,j'} k(y_j,y_{j'})
-\frac{2}{n^2}\sum_{i,j}k(x_i,y_j), \\
k(x,y)&=\exp\!\left(-\frac{\|x-y\|_2^2}{2\sigma^2}\right).
\end{aligned}
\label{eq:prompt_mmd}
\end{equation}
We choose $\sigma^2$ once per input-length analysis as the median squared
pairwise distance among 256 sampled embeddings from each of the six groups
(reference, internal baseline, and four interference groups). This gives
$\sigma^2=1.62008147$ at 2{,}048 tokens and $1.61263244$ at 512 tokens.
Within each analysis all comparisons share this bandwidth. We additionally
multiply $\sigma$ by 0.5 and 2. The main math-error $<$ code-error $<$ XSum
ordering is unchanged across all six length/bandwidth combinations.
The unbiased U-statistic, obtained by omitting within-group diagonal terms
and using $n(n-1)$ denominators, is respectively $-0.000010$, $0.001218$,
$0.084316$, $0.115059$, and $0.129737$ for the five groups in Figure~\ref{fig:domain_distance} at the default
2{,}048-token setting. Its slightly negative internal-baseline estimate is
consistent with finite-sample variation around zero.

\paragraph{Resampling and interpretation.}
We perform 2{,}000 prompt-level bootstrap resamples, holding the encoder and
bandwidth fixed and sharing the reference resampling weights across
comparisons. Raw 2.5th-97.5th percentile ranges are shown in
Figure~\ref{fig:domain_distance}. They are $[0.000996,0.001687]$ for the
internal baseline and $[0.001978,0.003220]$ for math-error. Both ranges lie above
their corresponding point estimates because of near-null resampling bias.
We therefore do not interpret these ranges as calibrated confidence intervals
or use their exclusion of zero to test equality. For code-error, controlled
APPS, and XSum the ranges are $[0.082520,0.089160]$,
$[0.113097,0.119811]$, and $[0.128525,0.133903]$, respectively.
The shared-reference bootstrap ranges for code-error minus math-error and
XSum minus code-error are $[0.080136,0.086475]$ and
$[0.042010,0.049054]$, respectively, supporting the observed separation
conditional on this representation. A separate pooled-label permutation
test uses 999 permutations and the $+1$ correction. These tests are
exploratory and not corrected for multiple comparisons. Neither resampling
procedure measures variation across training seeds or encoders.

\paragraph{Truncation, numerical verification, and Fisher comparison.}
At 2{,}048 tokens, the reference, internal baseline, and math-error samples
have no truncation, while code-error and controlled APPS each have two truncated
prompts and XSum has six. At 512 tokens the corresponding counts are
2, 1, 4, 233, 309, and 362 out of 1{,}024 per group. The ordering survives
this change, but this does not control all length or formatting effects.
The experiment uses torch 2.6.0 and transformers 4.57.6. An independent
NumPy CPU implementation
reproduces all 30 MMD$^2$ V- and U-statistic comparisons to absolute tolerance
$10^{-10}$ and verifies sample uniqueness and absence of exact reference
overlap.

The existing SFT/GRPO math code Fisher distances of 0.5952 and 0.7336
use APPS-first1024 prompts and each anchor's self-generated responses. Their
sample cohort differs from the code-error sample used here. They quantify
anchor-dependent diagonal importance profiles rather than fixed-encoder
prompt distributions. No math-error or XSum Fisher distance is inferred
from the text-domain results.

\section{The Ordering Is Stable across Corruption Types}
\label{app:experiment_corruption}
The source problems are from APPS+~\citep{Dou2024StepCoder}, the curated
APPS~\citep{Hendrycks2021APPS} dataset introduced with StepCoder. We construct
three trajectory conditions on these problems using
\texttt{deepseek-v4-flash}~\citep{DeepSeek2026V4}: correct (P), wrong format
(WF), and wrong reasoning (WR). The P/WF/WR conditions are constructed for
this study rather than supplied as part of the original APPS+ dataset.
Table~\ref{tab:apps} replaces sampled errors with controlled corruptions so that
the nature of the error can be attributed. The \GRPO\ anchor retains more under
all three variants: the margin over SFT is 14.65, 15.69 and 9.34 points for P,
WF and WR respectively. The margin is therefore not an artefact of \emph{how}
the trajectories were corrupted. At the same time, the three columns do not
separate surface format from reasoning content: the code benchmarks do not show
the divergence between WF and WR that the design intended to produce, and we
draw no conclusion from that contrast.

\begin{table}[htbp]
\centering\small
\setlength{\tabcolsep}{5pt}
\caption{Near-domain interference with controlled corruption of APPS+
trajectories: correct (P), wrong format (WF) and wrong reasoning (WR).
Accuracy (\%, $\uparrow$) and macro retention relative to each anchor ($\uparrow$).}
\label{tab:apps}
\begin{tabular}{@{}llccccc@{}}
\toprule
Anchor & Corruption & MATH500 & AIME24 & AIME25 & MacroRet & HumanEval+ / MBPP+ \\
\midrule
\GRPO & P  & \underline{81.60} & \textbf{18.33}    & \underline{16.88} & \underline{83.10} & 76.2 / 65.1 \\
\GRPO & WF & \textbf{82.55}    & \underline{17.71} & \textbf{17.29}    & \textbf{83.25}    & 75.6 / 63.2 \\
\GRPO & WR & 81.05             & 16.67             & 16.46             & 79.94             & 76.2 / 66.4 \\
SFT   & P  & 78.10             & 16.04             & 10.83             & 68.45             & 75.0 / 63.2 \\
SFT   & WF & 77.00             & 13.96             & 12.71             & 67.56             & 72.6 / 59.8 \\
SFT   & WR & 77.55             & 15.00             & 13.75             & 70.60             & 73.2 / 63.0 \\
\bottomrule
\end{tabular}
\end{table}

\section{Parameter Sensitivity: The Effect of Regularization Strength}
\label{app:experiment_regularisation}
Sweeping the EWC penalty over $\{0,10^{6},10^{7}\}$ reveals that it interacts
with the interference domain rather than helping uniformly. In the near-domain
block of Table~\ref{tab:main_results}, increasing $\lambda$ from $0$ to $10^{6}$ helps the
\GRPO\ anchor by 9.91 points and the SFT anchor by 3.25 points, widening the gap
between them. In the matched-domain block of the same table, the same change
helps \GRPO\ by 11.89 points and SFT by 8.20 points, leaving the two within a few
points of each other, and pushing $\lambda$ further to $10^{7}$ reverses the sign
of the comparison. We read this as evidence that a penalty computed on the
anchored capability protects it, but cannot create a difference between
objectives that the interference domain has already erased.

\section{Static Fisher Diagnostics}
\label{app:experiment_fisher}
\begin{figure}[htbp]
\centering
\includegraphics[width=\linewidth]{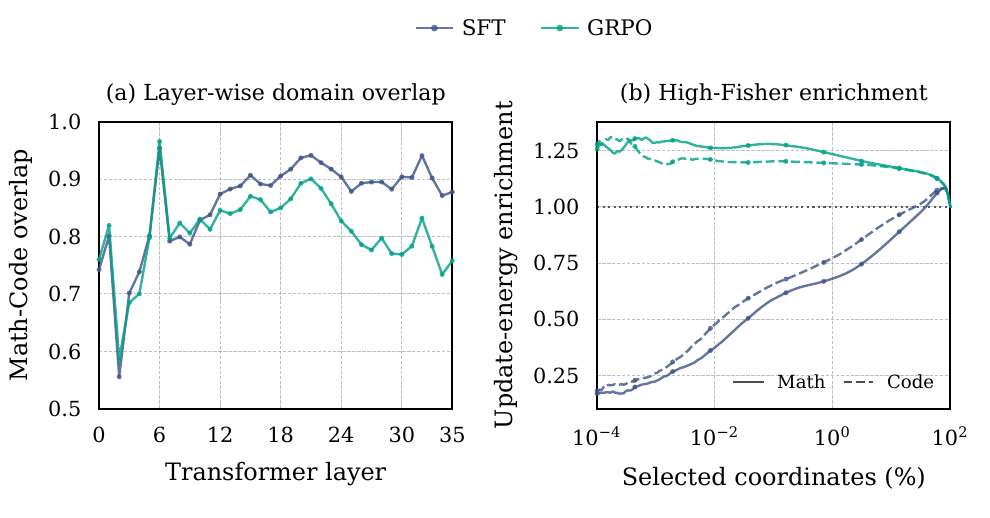}
\caption{Static Fisher diagnostics of the two anchors.
(a) Math Code importance-profile overlap, with each domain normalized
separately within each Transformer layer. Larger values indicate greater
overlap. (b) Base-to-anchor update-energy enrichment on globally ranked
high-Fisher coordinates. Solid and dashed lines use math and code rankings,
respectively. The dotted line denotes the uniform random-coordinate
expectation of one. Curves are descriptive and have no uncertainty bands.}
\label{fig:fisher_static}
\end{figure}

We examine the SFT step-592 and \GRPO\ step-410 anchors using their
math and code empirical diagonal Fisher estimates, each obtained from
1{,}024 self-generated responses. Figure~\ref{fig:fisher_static}
compares domain overlap within layers and the placement of the
base-to-anchor update on high-Fisher coordinates. Parameter scope and
preprocessing follow Appendix~\ref{app:geometry_details}.

\paragraph{Layer-wise domain overlap.}
For the parameter coordinates $G_l$ in layer $l$, we compute
\begin{equation}
O_{o,l}=\frac{\sum_{j\in G_l}\sqrt{F^{\mathrm{math}}_{o,j}F^{\mathrm{code}}_{o,j}}}
{\sqrt{\sum_{j\in G_l}F^{\mathrm{math}}_{o,j}}\,
 \sqrt{\sum_{j\in G_l}F^{\mathrm{code}}_{o,j}}},
\end{equation}
where $o$ denotes the anchor. Overlap is similar in early layers, but
\GRPO\ has lower overlap in every layer from 11 through 35.
The unweighted mean across layers 0-10 is 0.773 for SFT and 0.779 for
\GRPO. Across layers 24-35, the corresponding means are 0.895 for SFT and 0.785 for \GRPO.
Thus the math and code importance profiles are more separated in the
middle and later layers of the \GRPO\ anchor. This layer-normalized
comparison excludes the tied embedding/output tensor and final norm.
It does not decompose or identify the dominant source of the global
profile distance in Table~\ref{tab:geometry}.

\paragraph{Update-energy enrichment.}
Let $S_{o,c}(q)$ contain the top $q$ fraction of all coordinates ranked
by anchor $o$'s Fisher in domain $c$. We plot
\begin{equation}
E_{o,c}(q)=\frac{\sum_{j\in S_{o,c}(q)}(\Delta\theta_{o,j})^2}
{q\,\|\Delta\theta_o\|_2^2},
\qquad \Delta\theta_o=\theta_o-\theta_0.
\end{equation}
The random-coordinate expectation is one, up to coordinate-count rounding.
At all four exact budgets in Table~\ref{tab:fisher_projection}
(0.1\%, 1\%, 5\%, and 10\%), SFT is below this reference and \GRPO\ is
above it in both domains. At the top-1\% budget, enrichment is
0.680 for SFT and 1.234 for \GRPO\ under math rankings, and 0.771 and
1.194 under code rankings. The full curves show that SFT crosses above
one at larger budgets, so the ordering relative to the reference is
specific to the high-Fisher portion rather than all coordinate budgets.
Each curve is normalized by its own anchor's update energy and does not
compare absolute displacement magnitudes.

\paragraph{Numerical scope and interpretation.}
The continuous enrichment curves approximate global rankings with
131{,}072 bins over $\log_{10}F\in[-40,1]$, allocating update energy
proportionally within the boundary bin. The exact-budget results in
Table~\ref{tab:fisher_projection} are retained separately. These diagnostics
reuse the existing anchors and Fisher estimates. They are not independent
replications. Layer-wise overlap describes nonnegative importance profiles,
while enrichment describes the anchoring update, not a subsequent
interference step. The similar math and code enrichment curves within
each anchor do not establish domain-specific update allocation.
Because the Fisher estimates use each anchor's own generated responses,
the comparisons reflect both model and response differences. Neither
diagnostic alone establishes a causal explanation of retention. No
confidence interval or significance test is inferred from the curves.

\begin{table}[htbp]
\centering\small
\caption{Update energy share $100P_{o,c}(q)$ (\%) on the top-Fisher
coordinates. Columns specify the percentage of all coordinates selected.
Each row uses its anchor's own raw diagonal Fisher. The random-selection
expectation is approximately the selected coordinate percentage.}
\label{tab:fisher_projection}
\begin{tabular}{@{}llrrrr@{}}
\toprule
Anchor & Fisher domain & Top 0.1\% & Top 1\% & Top 5\% & Top 10\% \\
\midrule
SFT & Math & 0.0589 & 0.6801 & 3.9387 & 8.5737 \\
\GRPO & Math & 0.1280 & 1.2338 & 5.9659 & 11.7910 \\
SFT & Code & 0.0655 & 0.7714 & 4.4711 & 9.4545 \\
\GRPO & Code & 0.1204 & 1.1945 & 5.9237 & 11.7563 \\
\bottomrule
\end{tabular}
\end{table}

\section{Efficiency and Complexity}
\label{app:experiment_efficiency}
The anchors have approximately matched recorded token counts (939.21M versus
942.89M). The additional cost of the defense is a single extra pass
over 1{,}024 held-out samples for each Fisher diagonal, independent of how many
configurations reuse it, so EWC is cheap relative to interference training itself.
The spectral results also carry a storage implication: since reaching 90\% of the
update energy requires roughly 59\% as many directions for \GRPO\ as for SFT, a
low-rank approximation of a \GRPO\ update loses less capability per retained
direction, which is consistent with prior observations that RL updates are
compressible. Anchor-training and interference runtimes are reported in
Appendix~\ref{app:implementation}.

\section{Layer Freezing Gives Only a Transient Benefit}
\label{app:experiment_freezing}
\begin{figure}[htbp]
\centering
\includegraphics[width=0.92\linewidth]{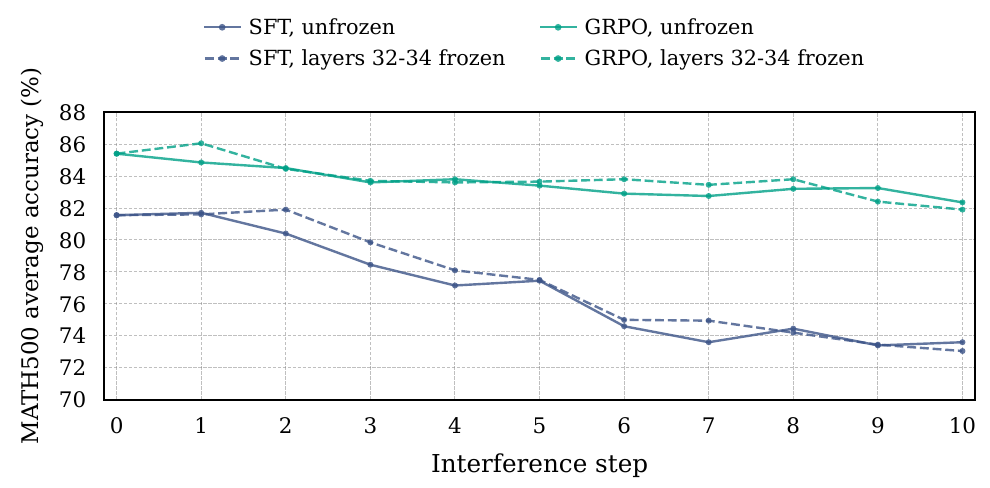}
\caption{MATH500 average accuracy over ten steps of APPS code-error continued
SFT, with and without layers 32-34 frozen. Step 0 reuses each anchor's
existing evaluation (SFT 81.55\%, GRPO 85.40\%), while steps 1-10 use the recorded
run evaluations. No uncertainty intervals are available.}
\label{fig:freeze}
\end{figure}

Figure~\ref{fig:freeze} compares continued SFT on the APPS code-error data,
with and without freezing all parameters in zero-based layers 32-34
(the 33rd-35th transformer blocks, including their norms). The launch
settings use learning rate $10^{-5}$, a constant schedule without warmup,
global batch size 128, ten steps, and EWC coefficient zero.
For each anchor, the frozen and unfrozen runs use the same training settings,
including training seed 42 and data order, except for layer freezing.
Freezing gives small improvements at some intermediate steps but no terminal
benefit: at step 10, SFT reaches 73.05\% with freezing versus 73.60\% without,
and GRPO reaches 81.90\% versus 82.35\%. These are exact recorded
MATH500 average accuracies over 500 problems with four responses each,
not values read from a plot or pass@4. Relative to the unfrozen runs, the
terminal changes are $-0.55$ percentage points for SFT and $-0.45$ for GRPO.
The terminal anchor ranking is unchanged. The full recorded trajectory is
provided in Appendix~\ref{app:freeze_results}.

\section{Behavior at Other Scales}
\label{app:experiment_scale}
Table~\ref{tab:scale17} reports a preliminary replication on Qwen3-1.7B. We
do not read retention differences from it, because the two anchors are not
comparable there: while \GRPO\ improves over its base model on all three
benchmarks, the SFT checkpoint is below the base model on MATH500 (47.30 against
55.55), so any comparison would attribute to objective what is actually a
difference in anchor quality. We additionally note that for this model the
interference trajectories originate from Qwen3-0.6B, which may itself be stronger
than the 1.7B anchors on these items. The corpus should be regenerated from the
anchor family before scale conclusions are drawn.

\begin{table}[htbp]
\centering\small
\caption{Preliminary replication on Qwen3-1.7B. Accuracy (\%, $\uparrow$).
The SFT anchor is below the base model on MATH500, so retention ratios are not
reported for this model.}
\label{tab:scale17}
\begin{tabular}{@{}llccc@{}}
\toprule
Setting & Condition & MATH500 & AIME24 & AIME25 \\
\midrule
Reference  & Base model            & 55.55 & 3.13 & 2.29 \\
Reference  & \GRPO\ anchor (step 444) & 67.05 & 9.38 & 4.17 \\
Reference  & SFT anchor (step 592)    & 47.30 & 3.96 & 2.92 \\
\midrule
After matched interference & \GRPO  & 65.15 & 7.50 & 2.92 \\
After matched interference & SFT    & 58.05 & 5.21 & 3.54 \\
After near interference    & \GRPO  & 66.70 & 7.71 & 4.17 \\
After near interference    & SFT    & 61.65 & 4.38 & 5.42 \\
\bottomrule
\end{tabular}
\end{table}

Table~\ref{tab:smollm} reports a replication on SmolLM3-3B. The two anchors are
not close in absolute strength here, the \GRPO\ anchor is above the SFT anchor on
all three benchmarks (77.75 versus 62.25 on MATH500). As on Qwen3-4B, the
retention gap is smallest under matched-domain interference (65.59\% versus
62.89\%) and larger under near- and far-domain interference (90.00\% versus
47.91\%, and 93.99\% versus 70.54\%). 

\begin{table}[htbp]
\centering\small
\caption{Replication on SmolLM3-3B. Accuracy (\%, $\uparrow$) and macro retention
relative to each anchor ($\uparrow$). }
\label{tab:smollm}
\begin{tabular}{@{}llcccc@{}}
\toprule
Setting & Condition & MATH500 & AIME24 & AIME25 & MacroRet \\
\midrule
Reference & Base model & 46.05 & 3.96 & 3.96 & - \\
Reference & \GRPO\ anchor & 77.75 & 15.83 & 14.58 & 100.00 \\
Reference & SFT anchor & 62.25 & 12.50 & 9.58 & 100.00 \\
\midrule
After matched interference & \GRPO & 66.20 & 5.00 & 11.67 & 65.59 \\
After matched interference & SFT & 58.05 & 4.58 & 5.63 & 62.89 \\
After near interference & \GRPO & 74.60 & 11.04 & 15.21 & 90.00 \\
After near interference & SFT & 39.35 & 3.54 & 5.00 & 47.91 \\
After far interference & \GRPO & 77.50 & 11.67 & 15.83 & 93.99 \\
After far interference & SFT & 59.70 & 8.75 & 4.38 & 70.54 \\
\bottomrule
\end{tabular}
\end{table}

\subsection{Qwen3-8B-Base: Matched, Near, and Far Interference}
\label{app:qwen3_8b}

We additionally evaluate Qwen3-8B-Base under matched-domain math-error,
near-domain code-error, and far-domain XSum interference. All six post-interference conditions use
\texttt{global\_step\_10}. The independent GRPO and SFT anchors are taken from
steps 393 and 592, respectively. Table~\ref{tab:qwen3_8b_accuracy} reports the
base model, both anchors, and their post-interference accuracies together with
anchor-relative macro retention. The SFT matched-domain endpoint uses two
generated responses per MATH500 question and eight per AIME24 or AIME25
question; its near-domain endpoint uses two and four, respectively. All other
conditions use four responses per MATH500 question and sixteen per AIME
question. Retention follows Equation~\ref{eq:ret}:
benchmark accuracies are divided by the corresponding anchor accuracies before
the three ratios are averaged with equal weight.

\begin{table}[htbp]
\centering\small
\setlength{\tabcolsep}{5pt}
\caption{Qwen3-8B-Base accuracy and macro retention (\%). All interference rows
use step 10. MacroRet is normalized by each model's own anchor, which has
100\% retention by definition. The base model has no retention value.}
\label{tab:qwen3_8b_accuracy}
\begin{tabular}{@{}lcccc@{}}
\toprule
Model / condition & MATH500 & AIME24 & AIME25 & MacroRet \\
\midrule
Base & 68.95 & 11.04 & 8.54 & -- \\
GRPO anchor (step 393) & 87.80 & 29.38 & 19.79 & 100.00 \\
SFT anchor (step 592) & 87.50 & 33.13 & 30.42 & 100.00 \\
\midrule
GRPO / Matched & 56.25 & 11.46 & 12.08 & 54.71 \\
SFT / Matched & 64.30 & 10.00 & 12.92 & 48.71 \\
GRPO / Near & 84.85 & 22.92 & 19.58 & 91.20 \\
SFT / Near & 81.70 & 26.67 & 21.67 & 81.70 \\
GRPO / Far & 87.75 & 28.75 & 21.46 & 102.08 \\
SFT / Far & 85.95 & 29.17 & 26.88 & 91.55 \\
\bottomrule
\end{tabular}
\end{table}

\begin{table}[htbp]
\centering\small
\setlength{\tabcolsep}{5pt}
\caption{Per-benchmark anchor-relative retention (\%) for Qwen3-8B-Base at
interference step 10. The final column reports $100-\mathrm{MacroRet}$.}
\label{tab:qwen3_8b_retention}
\begin{tabular}{@{}lccccc@{}}
\toprule
Model / condition & MATH500 & AIME24 & AIME25 & MacroRet & Loss \\
\midrule
GRPO / Matched & 64.07 & 39.01 & 61.05 & 54.71 & 45.29 \\
SFT / Matched & 73.49 & 30.19 & 42.47 & 48.71 & 51.29 \\
GRPO / Near & 96.64 & 78.01 & 98.95 & 91.20 & 8.80 \\
SFT / Near & 93.37 & 80.50 & 71.23 & 81.70 & 18.30 \\
GRPO / Far & 99.94 & 97.87 & 108.42 & 102.08 & -2.08 \\
SFT / Far & 98.23 & 88.05 & 88.36 & 91.55 & 8.45 \\
\bottomrule
\end{tabular}
\end{table}

\paragraph{Paired uncertainty and domain contrasts.}
We apply the paired question-bootstrap procedure in
Appendix~\ref{app:paired_bootstrap}. Within each benchmark, its original number of questions
is sampled with replacement, and the same sampled question indices are used
for both anchors and all six interference conditions. Benchmarks are sampled
independently, and all existing responses to each sampled question are kept
together. Each replicate recomputes the anchor denominators, benchmark
retention ratios, MacroRet, and GRPO--SFT gaps. The 95\% intervals use the
2.5th and 97.5th percentiles with linear interpolation, without clipping or
multiple-comparison adjustment.

\begin{table}[htbp]
\centering\small
\setlength{\tabcolsep}{4pt}
\caption{Qwen3-8B-Base macro retention (\%) and GRPO--SFT gaps (percentage
points) at interference step 10, with 95\% paired-bootstrap intervals.
The interactions compare near and far interference with matched interference.}
\label{tab:qwen3_8b_uncertainty}
\begin{tabular}{@{}lccc@{}}
\toprule
Condition & SFT & GRPO & GRPO $-$ SFT \\
\midrule
Matched & 48.71 [38.59, 57.65] & 54.71 [45.34, 61.63] & +6.00 [-2.32, 14.80] \\
Near & 81.70 [71.60, 91.64] & 91.20 [83.68, 98.53] & +9.50 [-3.31, 23.46] \\
Far & 91.55 [85.28, 97.63] & 102.08 [96.40, 108.82] & +10.53 [4.08, 18.67] \\
\midrule
Near $-$ Matched & \multicolumn{2}{c}{--} & +3.50 [-9.68, 16.04] \\
Far $-$ Matched & \multicolumn{2}{c}{--} & +4.54 [-4.35, 14.29] \\
\bottomrule
\end{tabular}
\end{table}

\section{Observations}
\label{app:experiment_observations}
Three conclusions follow from the tables above. First, whether one objective
yields a more stable capability depends on where the subsequent learning happens:
\GRPO-derived math ability has higher retention point estimates under near-
and far-domain interference. As shown in Table~\ref{tab:retention_uncertainty},
the paired interval excludes zero for the far-domain gap and the far-minus-matched
interaction, but includes zero for the near- and matched-domain gaps.
This supports a domain-dependent gap for the evaluated checkpoints.

\begin{figure}[htbp]
\centering
\includegraphics[width=\linewidth]{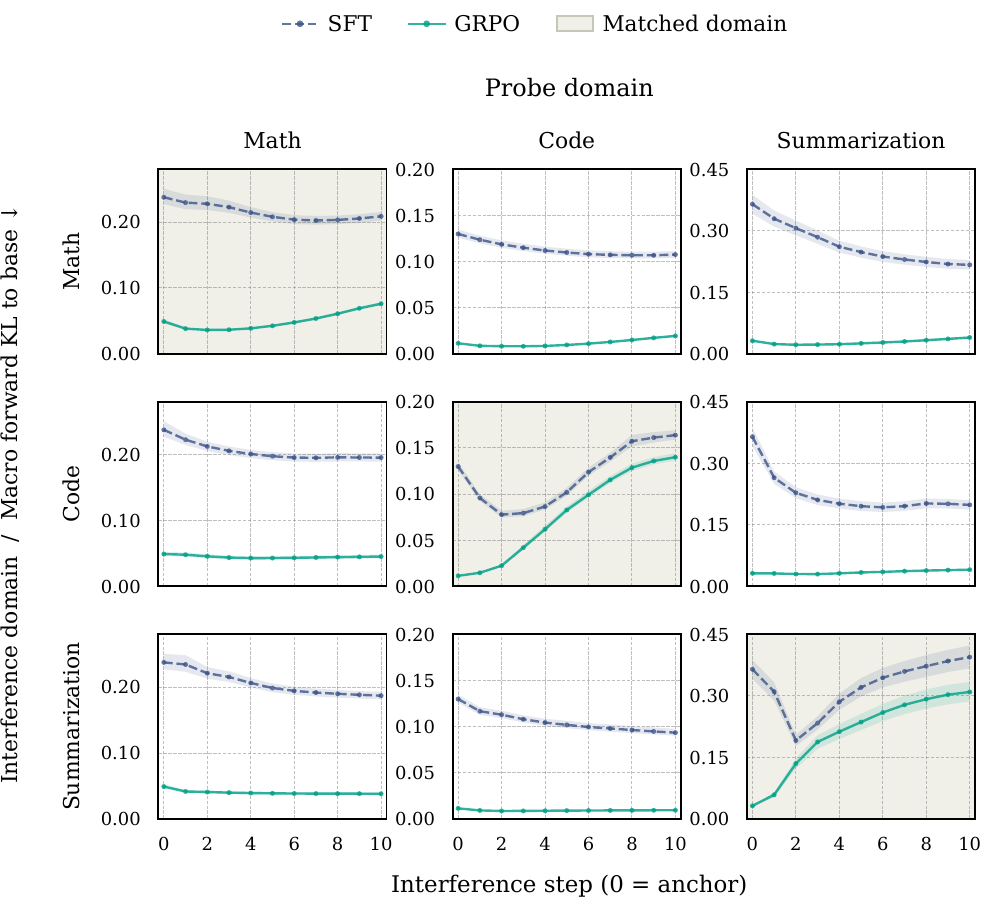}
\caption{Macro forward KL divergence to the base policy over interference
training steps, computed on 1{,}000 fixed contexts per probe domain ($\downarrow$
means less drift). Rows correspond to the interference domain and columns to the
probe domain. Shaded cells are those where the two coincide.}
\label{fig:kl}
\end{figure}

Second, initial anchor performance alone does not explain the retention pattern:
Table~\ref{tab:main_results} reports similar anchor accuracies but different
retention outcomes. KL drift alone also does not explain the retention pattern:
on matched probes, \GRPO\ moves more than SFT in Figure~\ref{fig:kl} while still
retaining comparably.

Third, the mechanism is geometric rather than algorithmic.
Table~\ref{tab:geometry} reports nearly orthogonal updates that differ in effective
rank and spectral concentration. Figure~\ref{fig:gradsim} characterizes whether
clean and corrupted gradients share early-layer support. Together, these
geometric properties make the \GRPO\ solution both easier to recover from a
compressed update and more dependent on the few directions it occupies, as
illustrated by the spectral interventions in Table~\ref{tab:spectral}.

\clearpage
\section{Distributional Drift Is Gated by Domain Match}
\label{app:kl_trajectories}

\end{document}